\documentclass{article} \usepackage{dualwm_arxiv,times}

\usepackage{amsmath,amsfonts,bm}

\def\eqref#1{equation~\ref{#1}}

\def\1{\bm{1}}

\DeclareMathAlphabet{\mathsfit}{\encodingdefault}{\sfdefault}{m}{sl}
\SetMathAlphabet{\mathsfit}{bold}{\encodingdefault}{\sfdefault}{bx}{n}

\usepackage{hyperref}
\usepackage{url}
\usepackage{graphicx}
\usepackage{xcolor}
\usepackage{booktabs}
\usepackage{multirow}
\usepackage{amssymb}
\usepackage{placeins}
\usepackage{capt-of}
\usepackage{wrapfig}
\usepackage{needspace}

\title{Beyond a Single Latent Space: A Dual-Latent World Model for Long-Horizon Planning}

\author{\normalfont
Delin Zhao\textsuperscript{1,2}, Zhengrong Yue\textsuperscript{4}, Shaobin Zhuang\textsuperscript{4}, Junlin He\textsuperscript{2}, Xiaoyu Chen\textsuperscript{3},\\
Zikang Wang\textsuperscript{4}, Yuxin Liu\textsuperscript{3}, Limin Wang\textsuperscript{1}, Yali Wang\textsuperscript{3,*}\\[0.6em]
\small \textsuperscript{1}Nanjing University\\
\small \textsuperscript{2}Shenzhen University of Advanced Technology\\
\small \textsuperscript{3}Shenzhen Institutes of Advanced Technology, Chinese Academy of Sciences\\
\small \textsuperscript{4}Shanghai Jiao Tong University\\[0.4em]
\small \textsuperscript{*}Corresponding author.}
\date{}

\newcommand{\dlwm}{Dual-WM}
\newcommand{\lore}{\textsc{LoRe}}

\hypersetup{
  pdftitle={Beyond a Single Latent Space: A Dual-Latent World Model for Long-Horizon Planning},
  pdfauthor={Delin Zhao, Zhengrong Yue, Shaobin Zhuang, Junlin He, Xiaoyu Chen, Zikang Wang, Yuxin Liu, Limin Wang, Yali Wang},
  pdfsubject={Dual-Latent World Model for Long-Horizon Planning}
}
\begin{document}

\maketitle

\begin{abstract}

Latent world models often struggle with long-horizon planning, even when their short-term predictions are accurate. Prediction errors accumulate during recursive rollout, while distance concentration in high-dimensional latent spaces can weaken the distinction between states at different distances from a goal. We introduce the Dual-Latent World Model (Dual-WM), which separates local execution from long-range planning through two distinct state representations and dynamics models. The low-level model captures fine-grained action-conditioned transitions, while the high-level model uses learned macro-actions to plan over longer temporal spans. We further propose Long-Horizon Representation Learning with Weighted Rollout (LoRe), which supervises self-generated predictions at both levels. Motivated by an analysis of recursive error propagation, LoRe uses exponential horizon weights with separate decay rates to balance multi-step supervision at each temporal scale. During planning, the high-level model generates latent subgoals, and the low-level model refines them into actions for precise execution. We evaluate from-scratch Dual-WM on five goal-conditioned visual control tasks against the task-wise strongest baselines without actor-guided proposals. At goal offsets of 50 and 100 environment steps, mean success increases from 75.9\% to 84.4\% and from 61.4\% to 69.5\%, respectively. At offset 100, Dual-WM outperforms these baselines on all five tasks and improves mean success over LeWM by 30.8 percentage points. Ablations and supporting analyses provide evidence that the proposed design learns more informative state representations for goal evaluation and improves consistency under recursive prediction. These results highlight the value of separating temporal roles and training across multiple horizons for reliable latent planning.
Our core implementation is available at \url{https://github.com/DeLin1001/Dual-WM-Official}.

\end{abstract}

\setlength{\parskip}{4.5pt}

\section{Introduction}
\label{sec:intro}

Latent world models provide a compact space in which an agent can predict the
consequences of its actions
\citep{hafner2019planet,hafner2023dreamerv3,hansen2023tdmpc2} and evaluate
progress toward a goal. These two uses impose distinct demands on the
representation: it must support accurate action-conditioned transitions, while
its distances must remain informative for ranking candidate futures. Local
predictive accuracy, however, does not guarantee consistent rollouts or
informative goal distances over long horizons.

\begin{figure}[t]
    \centering
    \includegraphics[width=\linewidth]{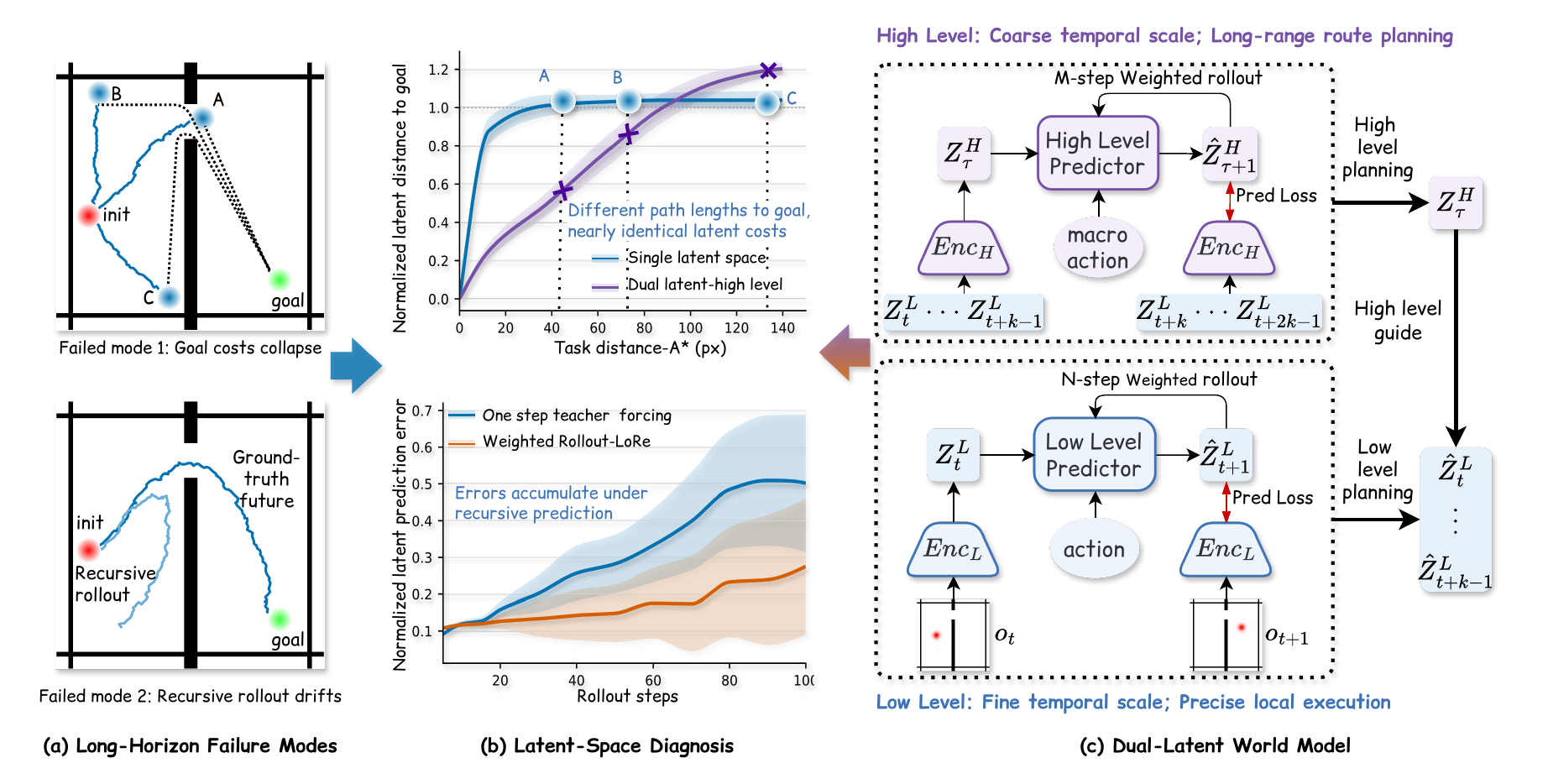}
    \caption{Overview of the long-horizon planning problem and our dual-latent
    solution. (a) Two failure modes in a single latent space: goal costs
    collapse for distant states, and recursive rollouts drift from the
    ground-truth future. (b) Latent-space diagnostics: the high-level space
    preserves task-distance contrast, while weighted rollout training reduces
    multi-step prediction error relative to one-step teacher forcing. (c)
    \dlwm{} assigns coarse route planning to a high-level latent and precise
    local execution to a low-level latent.}
    \label{fig:overview}
\end{figure}

This gap appears in two forms. First, repeatedly composing a locally trained transition
model can accumulate prediction errors, causing imagined trajectories to diverge
from their true evolution \citep{chen2026compounding,du2026vlwm,huo2026flowjepa}.
Second, even accurately predicted candidate futures can be difficult to rank
\citep{wang2026dalewm,hu2026scale}. In high-dimensional, isotropically regularized
representations \citep{maes2026leworldmodel,sobal2025pldm,zhao2026subjepa}, latent
distances can concentrate around a dimension-dependent baseline as temporal
correlations decay \citep{aggarwal2001distance,thil2026sdjepa,wu2026viscore},
leaving distant candidates with nearly indistinguishable goal costs. One-step
prediction constrains transitions between neighboring states, but does not
directly require distances between remote states to reflect task progress.
Long-horizon planning therefore
requires both predictive consistency and a geometry that supports meaningful
comparisons beyond local neighborhoods.
Without both, a planner may either imagine the wrong future or mis-rank otherwise
accurate candidate futures.
Appendix~\ref{app:ranking-margin} relates these requirements through a simple
condition for preserving candidate rankings.
Figure~\ref{fig:overview} illustrates both failure modes on held-out TwoRoom data.

Existing approaches address these challenges through improvements to predictive
accuracy, search efficiency, and temporal abstraction. Flat world models
strengthen prediction
\citep{du2026vlwm,gao2026fastlewm,huo2026flowjepa,liu2026odeworld,rakhimov2026qantara}
or introduce proposals and subgoals to facilitate search
\citep{cheng2026sage,wang2026prism,sun2026intact,huang2026leflow}. Hierarchical
models shorten the effective rollout depth by predicting temporally extended
transitions
\citep{zhang2026hwm,cabeza2026ffjepa,liu2026proworld,barbeau2026lago,gumbsch2024thick}.
However, in approaches that retain a shared latent space
\citep{caselli2026hilewm,thil2026sdjepa}, both local execution and long-range goal
evaluation still depend on the same state geometry; shortening the rollout leaves
this geometry unchanged, so when distances saturate, distant candidates remain
nearly indistinguishable.

The underlying issue is that the same state geometry must support two different
temporal roles. Precise local control requires sensitivity to small state changes
that affect immediate action outcomes \citep{zhang2026deltajepa,shi2026coco}.
Long-range planning instead requires comparisons that are not dominated by locally
important variations and that remain informative about progress toward distant
goals \citep{bai2026tdjepa,hu2026scale}. A shared representation must accommodate
both demands within the same geometry, although local prediction objectives do not
directly enforce long-range discrimination. We therefore separate representations
by temporal role, allowing each space to organize state differences according to
the transitions and planning decisions it supports.

We introduce Dual-Latent World Model (\dlwm{}), which couples a low-level
representation and dynamics model for primitive actions with a distinct
high-level representation and dynamics model for learned macro-actions.
Macro-Action Prior Shaping (MAPS) regularizes the macro-action posterior toward a
standard Gaussian prior to support sampling and optimization during planning.
We introduce \lore{} (Long-Horizon Representation Learning with Weighted Rollout)
to improve recursive consistency and constrain state relations beyond one-step
neighborhoods. \lore{} supervises self-generated rollouts at both scales using
separately parameterized horizon weights. At inference time, \dlwm{} first plans
high-level latent subgoals, then refines them into primitive actions, and finally
switches to direct low-level planning for precise goal convergence.

Our main contributions are threefold:

\begin{itemize}
  \item We introduce \textbf{\dlwm{}}, a world model that separates state
  representations and dynamics by temporal role. Macro-Action Prior Shaping
  (MAPS) regularizes learned macro-actions for sampling-based search, while
  hierarchical planning connects long-range guidance to precise local execution.

  \item We propose \textbf{\lore{}}, a training mechanism that supervises
  self-generated rollouts at both temporal scales with separately parameterized
  horizon weights, extending predictive supervision beyond one-step transitions.

  \item We evaluate \dlwm{} on five goal-conditioned visual planning tasks
  spanning continuous control and discrete-action, game-like planning, with
  comparisons against flat and hierarchical world models and diagnostics of
  the learned representations.
\end{itemize}

\section{Related Work}
\label{sec:related}

\paragraph{Latent prediction and control.}
Latent dynamics support planning from pixels and policy learning in imagined
trajectories \citep{hafner2019planet,hafner2023dreamerv3,hansen2023tdmpc2}.
Our setting is reconstruction-free, goal-conditioned planning:
DINO-WM predicts pretrained visual features \citep{zhou2024dinowm},
PLDM learns from reward-free trajectories \citep{sobal2025pldm}, and
LeWM jointly learns an encoder and predictor with anti-collapse
regularization \citep{maes2026leworldmodel}. Beyond one-step prediction,
VLWM, Fast-LeWM, and Flow-JEPA change how extended futures are predicted
\citep{du2026vlwm,gao2026fastlewm,huo2026flowjepa}. LoRe instead retains
recursive dynamics and supervises their self-generated rollouts at both
temporal scales, with separate decayed horizon weights.

\paragraph{Geometry and temporal abstraction.}
Planning-oriented objectives reshape representations through action
alignment, physical-state distances, reachability, or temporal ordering
\citep{wang2026dalewm,hu2026scale,li2026rcaux,bai2026tdjepa}.
SD-JEPA separates progression and content within an embedding
\citep{thil2026sdjepa}. Dual-WM separates dynamical roles: each state
space has its own predictor and action timescale, and the high-level
encoder maps predicted low-level windows into the subgoal space.
HWM and Hi-LeWM are close hierarchical precedents
\citep{zhang2026hwm,caselli2026hilewm}; both retain a shared latent-state
interface across scales. Our learned cross-space interface allows
macro-action subgoals and final primitive-action control to use different
representations. HWM's reported autoregressive objective uses uniform
rollout weights, while RC-aux permits horizon-weighted open-loop
supervision \citep{li2026rcaux}. LoRe uses scale-specific exponential decay
motivated by recursive error propagation.
Appendix~\ref{app:related} and Table~\ref{tab:positioning} detail these
connections and complementary approaches to action search.

\section{Method}
\label{sec:method}

\begin{figure}[t]
  \centering
  \includegraphics[width=\linewidth]{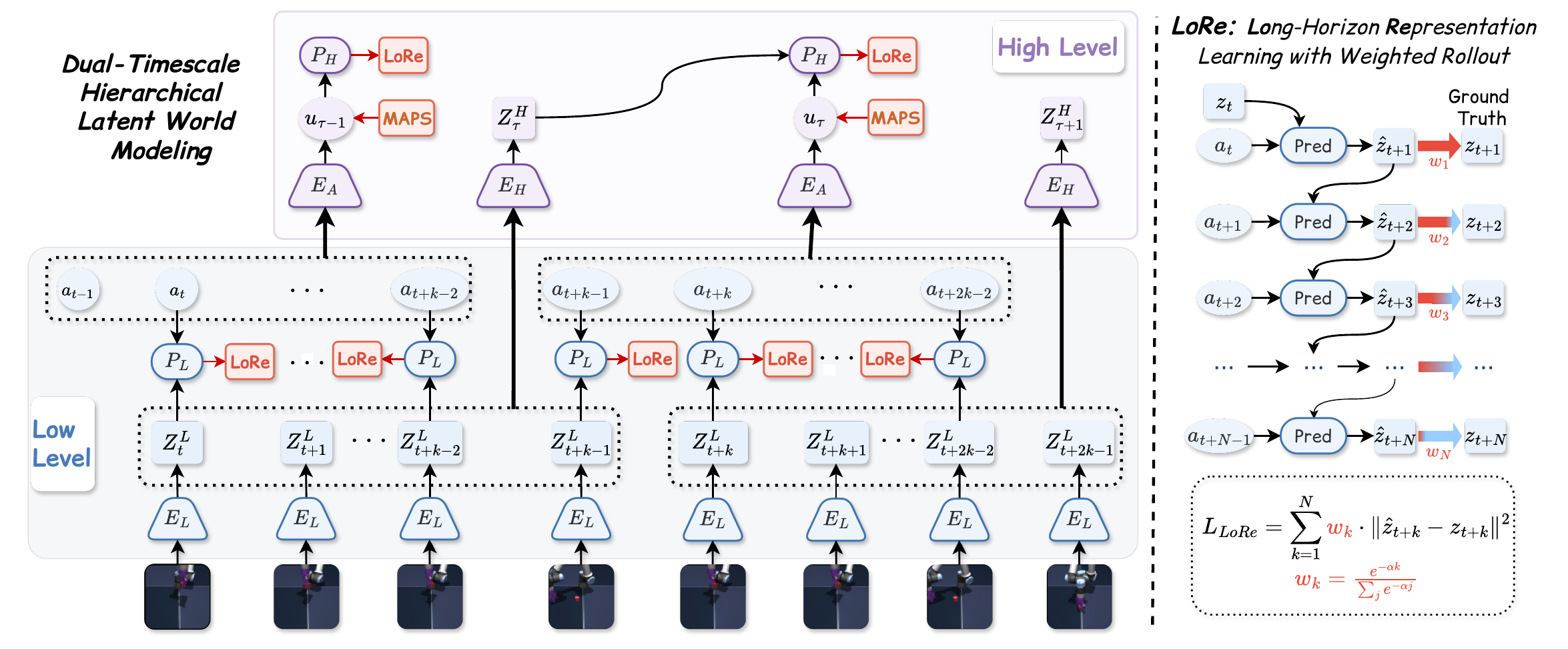}
  \caption{Training the dual-timescale latent world model. The low-level
  encoder $E_L$ and predictor $P_L$ model fine-scale transitions, while $E_H$
  encodes length-$k$ low-level latent windows and $P_H$ models temporally
  extended transitions. The action encoder $E_A$
  produces macro-actions, MAPS regularizes their posterior, and \lore{}
  supervises self-generated predictions at both levels.}
  \label{fig:training-framework}
\end{figure}

\begin{figure}[t]
  \centering
  \includegraphics[width=\linewidth]{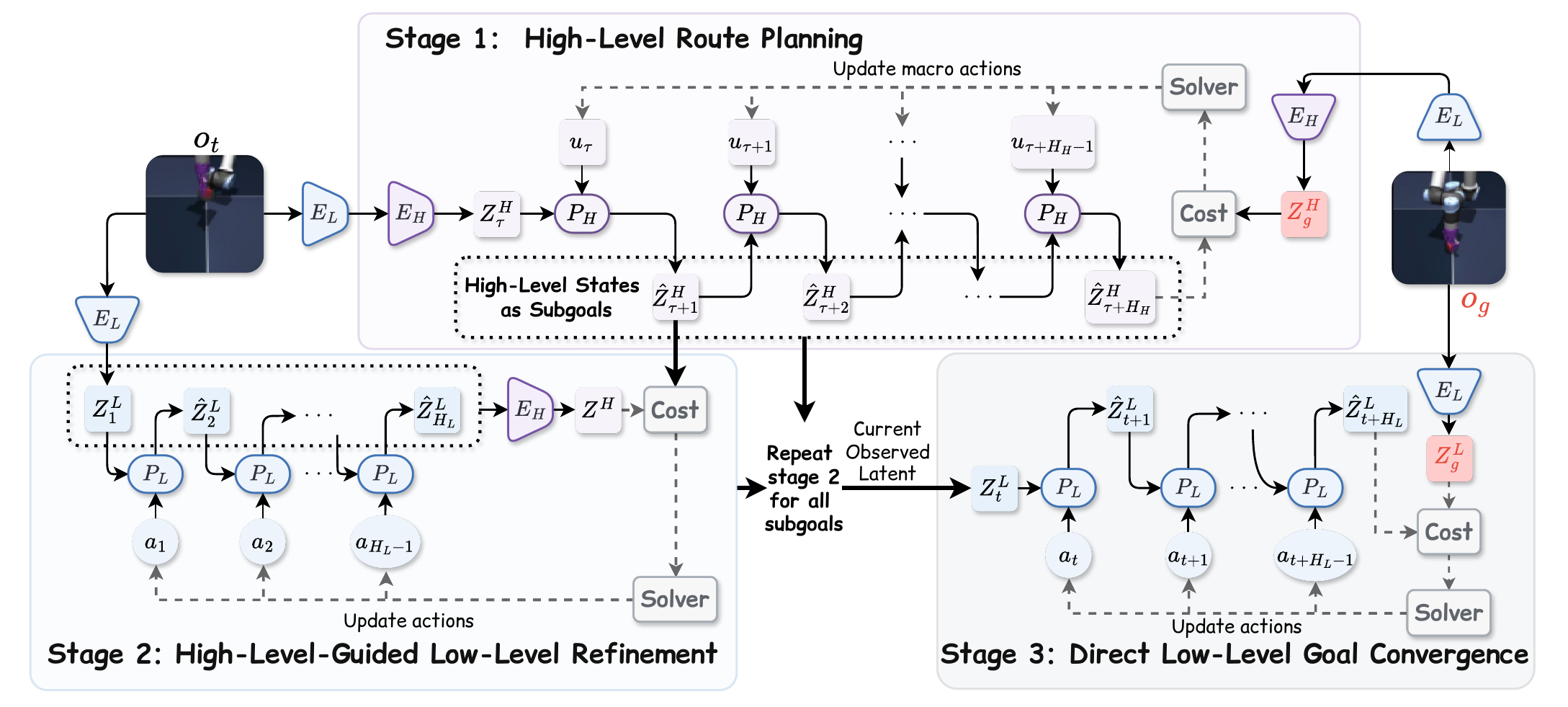}
  \caption{Three-stage planning with the trained dual-latent world model.
  Stage 1 plans a high-level route and produces latent subgoals; Stage 2
  refines subgoals with low-level actions while comparing latent windows through $E_H$;
  Stage 3 performs direct low-level goal convergence.}
  \label{fig:planning-framework}
\end{figure}

\dlwm{} assigns local execution and long-range planning to different state
spaces, then couples them through a learned projection used during subgoal
refinement. Figures~\ref{fig:training-framework} and
\ref{fig:planning-framework} show training and three-stage planning.
We first define the two dynamics models and their macro-action interface,
then describe \lore{} and the planner that uses them.

\subsection{Problem setting and notation}
\label{sec:setting}
We learn from an offline dataset $\mathcal D$ of observation--action
trajectories. In the model-level notation
$(o_0,a_0,o_1,\ldots,a_{T-1},o_T)$, $t$ indexes low-level model steps.
At deployment, the agent receives the current observation $o_t$ and a goal
observation $o_g$, and selects primitive actions by planning with the learned
models. No task reward or ground-truth state-distance target is required by
the predictive objectives below. Each low-level model input $a_t$ represents
$f$ consecutive environment actions, so one low-level model step covers $f$
environment steps. We write a macro-action segment as
$A_t=(a_t,\ldots,a_{t+k-1})$, where $k$ is its span in low-level model
steps and the number of low-level latents in each high-level input window.
Thus, one high-level transition covers $k$ low-level model steps, or $kf$
environment steps.
In the diagrams, $\tau$ indexes high-level transitions: advancing from
$\tau$ to $\tau+1$ corresponds to advancing from $t$ to $t+k$ in the
low-level time index.
For clarity, we describe fixed $k$ within a model configuration.
We use $N$ and $M$ for the low- and high-level training rollout lengths,
respectively, and $H_L,H_H$ for their planning horizons. We suppress
any additional context-window arguments in the predictor notation.

\subsection{Dual-latent dynamics and macro-actions}
\label{sec:dual}
Let $W_t^L$ denote the length-$k$ window of low-level latents associated
with high-level state $t$. When only one frame is available, its low-level
latent is repeated to fill this window. The observation encoder $E_L$ and high-level
state encoder $E_H$ define
\begin{equation}
  z_t^L=E_L(o_t)\in\mathbb R^{d_L},
  \qquad
  z_t^H=E_H(W_t^L)\in\mathbb R^{d_H}.
  \label{eq:dual-encoding}
\end{equation}
The two spaces have separate dynamics predictors:
\begin{equation}
  \hat z_{t+1}^L=P_L(z_t^L,a_t),
  \qquad
  \hat z_{t+k}^H=P_H(z_t^H,u_t).
  \label{eq:dual-dynamics}
\end{equation}
Here $u_t\in\mathbb R^{d_u}$ summarizes $A_t$ through a posterior
$q_\phi(u_t\mid A_t)$ parameterized by the macro-action encoder $E_A$.
Low-level prediction preserves distinctions needed for primitive action
outcomes. High-level prediction is trained against states separated by $k$
low-level model steps, allowing its geometry to emphasize distinctions relevant at that
temporal scale. Both levels share the visual backbone, with $E_H$ mapping
low-level windows into the state space used for high-level prediction and
goal evaluation.

During training, the high-level branch receives
$E_H(\operatorname{sg}(W_t^L))$, where $\operatorname{sg}$ preserves its
argument's value and stops its gradient. High-level losses therefore update
the high-level modules without changing $E_L$ or $P_L$.
Low-level learning retains its fine-scale predictive objective, while $E_H$
learns how to organize the supplied features for macro-action dynamics.

\paragraph{Macro-action prior shaping.}
A macro-action space learned only on encoded dataset segments need not be
well behaved under unconstrained search. MAPS adds
\begin{equation}
  \mathcal L_{\mathrm{MAPS}}
  =\mathbb E_{A_t\sim\mathcal D}
  D_{\mathrm{KL}}\!\left(q_\phi(u_t\mid A_t)
  \,\|\,\mathcal N(0,I_{d_u})\right).
  \label{eq:maps}
\end{equation}
The high-level prediction loss encourages $u_t$ to retain information about
the segment's effect, while the KL penalty gives candidate generation a
common prior. At planning time, macro-actions are optimization variables;
their posterior encoder is needed for training, not for encoding unknown
future actions. MAPS therefore regularizes the macro-action interface for
sampling-based planning, complementing the state representations used for
goal evaluation.

\subsection{Horizon-weighted self-generated rollouts}
\label{sec:lore}
\lore{} trains each dynamics model on the recursive predictions it will use
during planning, while controlling how different rollout horizons contribute
to learning. The two scales use separately parameterized weights because
low-level and high-level prediction operate over different temporal
spans and can exhibit different error-amplification rates.
Let $s\in\{L,H\}$ index the temporal scale, with
$\Delta_L=1$ and $\Delta_H=k$, measured in low-level model steps.
For a common loss expression, let
$n_L=N$ and $n_H=M$. Write $b_t^L=a_t$ and $b_t^H=u_t$.
Starting from an encoded anchor, we generate
\begin{equation}
  \hat z_{t\mid t}^{s}=z_t^s,\qquad
  \hat z_{t+h\Delta_s\mid t}^{s}
  =P_s\!\left(\hat z_{t+(h-1)\Delta_s\mid t}^{s},
              b_{t+(h-1)\Delta_s}^{s}\right).
  \label{eq:recursive-training}
\end{equation}
The low-level rollout uses the recorded low-level action inputs. The high-level
rollout uses macro-actions encoded from the corresponding recorded segments.
After initialization, neither rollout is reset to an observed future state.
Future observations supply targets $z_{t+h\Delta_s}^s$:
\begin{equation}
  \mathcal L_{\mathrm{LoRe}}^s
  =\mathbb E_{\mathcal D,q_\phi}
    \sum_{h=1}^{n_s} w_h^s
    \left\|\hat z_{t+h\Delta_s\mid t}^{s}
                 -z_{t+h\Delta_s}^{s}\right\|_2^2,
  \qquad
  w_h^s=\frac{\exp(-\alpha_s h)}
              {\sum_{j=1}^{n_s}\exp(-\alpha_s j)}.
  \label{eq:lore}
\end{equation}
The expectation over $q_\phi$ applies to the high-level rollout.
The loss differentiates through the recursive prediction chain and the
encoded future targets; no additional stop-gradient is applied to those
targets. The branch boundary above still prevents high-level target and
prediction losses from updating the low-level encoder.
Each scale has its own horizon $n_s$ and decay $\alpha_s\geq0$:
one high-level step covers $k$ low-level model steps, or $kf$ environment
steps, so equal step indices do not represent equal physical horizons.
Uniform weighting is recovered at
$\alpha_s=0$, and finite decay retains supervision at every included horizon.

\paragraph{From recursive error to horizon weights.}
Following \citet{asadi2018lipschitz}, fix a control sequence and a shared
encoded starting state. Suppose $P_s$ is $L_s$-Lipschitz on the encoded and
predicted inputs encountered, with one-step residual at most $\epsilon_s$.
For $e_h^s=\|\hat z_{t+h\Delta_s\mid t}^{s}-z_{t+h\Delta_s}^{s}\|_2$
and $e_0^s=0$, the prediction error satisfies
\begin{equation}
  e_{h+1}^s\leq L_s e_h^s+\epsilon_s,
  \qquad
  e_h^s\leq\epsilon_s S_h(L_s),\qquad
  S_h(L)=\sum_{j=0}^{h-1}L^j.
  \label{eq:error-propagation}
\end{equation}
The reference weights $v_h^s\propto S_h(L_s)^{-2}$ equalize weighted
squared-error bounds. When $L_s>1$, their long-horizon behavior approaches
geometric decay, motivating the exponential family in Eq.~(\ref{eq:lore}).
Finite decay tempers amplified distant errors while retaining multi-step
supervision; separate decays accommodate the two scales' different error
growth. Appendix~\ref{app:rollout-weighting} gives the finite-horizon derivation
and its extension to history-conditioned predictors.

The full training objective combines predictive supervision,
macro-action prior shaping, and state regularization:
\begin{equation}
  \mathcal L
  =\lambda_L\mathcal L_{\mathrm{LoRe}}^L
   +\lambda_H\mathcal L_{\mathrm{LoRe}}^H
   +\beta\mathcal L_{\mathrm{MAPS}}
   +\mathcal L_{\mathrm{reg}}.
  \label{eq:training-objective}
\end{equation}
Here $\mathcal L_{\mathrm{reg}}$ is the isotropic Gaussian state regularizer
for anti-collapse, following LeWM~\citep{maes2026leworldmodel}, with the
regularization weights at both levels included in this term.
The $h=1$ terms provide one-step supervision within the LoRe losses.

\subsection{Coarse-to-fine planning across the two spaces}
\label{sec:planning}
Let $\Phi_s^{(m)}(z,B)$ denote $m$ recursive applications of $P_s$ starting
from $z$ under an action sequence $B$.
The planning horizons $H_L,H_H$ need not equal the training horizons
$N,M$. Encode the current and goal observations in the low-level space,
and their corresponding latent windows with $E_H$ to obtain the
high-level representations.

\paragraph{Stage 1: High-level route planning.}
We optimize a macro-action sequence $U=(u_0,\ldots,u_{H_H-1})$ using
\begin{equation}
  U^\star\approx\arg\min_U
  \left\|\Phi_H^{(H_H)}(z_t^H,U)-z_g^H\right\|_2^2.
  \label{eq:high-planning}
\end{equation}
CEM generates and updates candidates in the continuous macro-action space
regularized by MAPS.
Intermediate predictions under $U^\star$ provide a sequence of high-level
subgoals $\tilde z_1^H,\ldots,\tilde z_{H_H}^H$.
A route spanning $kH_H$ low-level model steps, or $kfH_H$ environment steps,
requires $H_H$ high-level transitions, reducing the prediction depth of
the route search before low-level refinement.

\paragraph{Stage 2: High-level-guided low-level refinement.}
Given the selected subgoal $\tilde z_j^H$, let $\widehat W_t^L(A)$ denote
the length-$k$ latent window constructed from the low-level rollout for
comparison with that subgoal. We optimize a sequence of low-level action inputs
$A=(a_0,\ldots,a_{H_L-1})$ through
\begin{equation}
  A^\star\approx\arg\min_A
  \left\|E_H\!\left(\widehat W_t^L(A)\right)
                  -\tilde z_j^H\right\|_2^2.
  \label{eq:subgoal-refinement}
\end{equation}
The dynamics remain low-level, but the comparison is high-level.
This projection is the coupling between the two state spaces: the low-level
plan is evaluated by how its latent window matches the abstract subgoal, without
requiring an inverse map from $z^H$ to $z^L$ or a decoder from macro-actions
to primitive actions. Execution is grounded again in the new observation,
and refinement proceeds toward successive subgoals.

\paragraph{Stage 3: Direct low-level goal convergence.}
After a fixed number of high-level-guided refinement rounds, the planner
switches to direct low-level goal convergence, replacing the projected
subgoal cost by
\begin{equation}
  A^\star\approx\arg\min_A
  \left\|\Phi_L^{(H_L)}(z_t^L,A)-z_g^L\right\|_2^2.
  \label{eq:goal-convergence}
\end{equation}
States that match in the high-level space can still differ in details that
matter for completion. Direct low-level goal matching retains sensitivity to
those differences. The three stages thus use high-level dynamics to choose
a route, low-level dynamics with projected costs to execute it, and low-level
costs to finish precisely.

\section{Experiments}
\label{sec:experiments}

Our experiments first assess (Q1) whether Dual-WM improves long-horizon
goal reaching across five visual control tasks. We then examine the two
challenges motivating our design: (Q2) whether latent goal distances
distinguish progress over long ranges, and (Q3) whether recursive
predictions retain task-relevant physical-state information. Finally,
controlled ablations assess (Q4) the contributions of the learned
high-level representation, rollout training, and macro-action prior shaping.

\subsection{Experimental setup}
\label{sec:experimental-setup}
We evaluate visual, goal-conditioned planning on TwoRoom, Reacher,
Sokoban-Long, PushT, and Cube-Single (Figure~\ref{fig:envs}). Our evaluation
includes Sokoban-Long, a game-like environment with discrete actions, to assess
whether the benefits of dual-latent planning extend beyond continuous-control
tasks. Goal-reaching
success is reported at the displayed offsets. Here, a goal offset $m$ denotes
$m$ original environment steps between the initial and goal observations in
the dataset. Our trained and reproduced entries in the full planning tables
are evaluated with three planning seeds ($0$, $1$, and $42$), using $200$
goal-reaching tasks per task and offset for each seed. Each seed selects a
new task set, shared across methods with identical episode IDs, start states,
and goals, and also fixes the planner's search randomness. For each task and
offset, all methods evaluated by us use identical environment-step budgets,
success criteria, and early-termination rules. Entries shown with $\pm$
report the mean and standard deviation across these seeded evaluations,
which vary both task sampling and search randomness
(Appendix~\ref{app:baselines}); quoted results retain the reporting
convention of the original publication. PushT
additionally includes offset $75$ because it is the long-horizon setting
reported for HWM. The
baselines are LeWM~\citep{maes2026leworldmodel}, DINO-WM~\citep{zhou2024dinowm},
PLDM~\citep{sobal2025pldm}, HWM~\citep{zhang2026hwm},
Hi-LeWM~\citep{caselli2026hilewm}, Fast-LeWM~\citep{gao2026fastlewm},
RC-aux~\citep{li2026rcaux}, two INTACT variants~\citep{sun2026intact},
JEPA-WM~\citep{terver2025whatdrives}, VLWM~\citep{du2026vlwm},
and our Gemini 3.8 Flash closed-loop controller.
Table~\ref{tab:aggregate} summarizes the extended-offset comparison.
Appendix~\ref{app:full-results} reports all task-level results, and
Appendix~\ref{app:baselines} distinguishes the baseline settings.
Training and planning configurations are in Appendix~\ref{app:configuration}.

\subsection{Q1: Does Dual-WM improve long-horizon planning?}
\label{sec:planning-performance}

Table~\ref{tab:aggregate} compares the same from-scratch Dual-WM variant
against LeWM and the strongest non-Dual-WM result without actor-guided
proposals on each task.
INTACT uses Pure CEM in this comparison; Actor+CEM is compared separately
with actor-guided Dual-WM in Table~\ref{tab:actor-comparison}.
Full results and protocols are in Appendices~\ref{app:full-results}
and~\ref{app:baselines}. The task-wise aggregate selects the strongest
eligible result from flat models, hierarchical models, and the external
controller for each task and offset.

\begin{table}[t]
  \centering
  \caption{Success (\%) at goal offsets $50$ and $100$ environment steps.
  Dual-WM uses the from-scratch variant throughout. \emph{Best other}
  selects the highest non-Dual-WM mean without actor-guided proposals
  separately for each task and offset, with the selected method shown beside
  each value. The external controller is included. Bold marks the largest
  displayed mean per task/offset. Full results and uncertainty are in
  Tables~\ref{tab:results-tworoom}--\ref{tab:results-cube}.}
  \label{tab:aggregate}
  \small
  \setlength{\tabcolsep}{0pt}
  \renewcommand{\arraystretch}{1.18}
  \begin{tabular*}{\linewidth}{@{\extracolsep{\fill}}lrr@{\hspace{3pt}\extracolsep{0pt}}l@{\extracolsep{\fill}}rrr@{\hspace{3pt}\extracolsep{0pt}}l@{\extracolsep{\fill}}r@{}}
    \toprule
    & \multicolumn{4}{c}{Offset 50} & \multicolumn{4}{c}{Offset 100} \\
    \cmidrule(lr){2-5}\cmidrule(lr){6-9}
    Task & LeWM & \multicolumn{2}{c}{Best other} & Dual-WM & LeWM & \multicolumn{2}{c}{Best other} & Dual-WM \\
    \midrule
    TwoRoom & 61.50 & 94.83 & {\scriptsize\textcolor{black!65}{(Gemini)}} & \textbf{99.33} & 27.00 & 90.00 & {\scriptsize\textcolor{black!65}{(Gemini)}} & \textbf{92.50} \\
    Reacher & 85.33 & 92.33 & {\scriptsize\textcolor{black!65}{(INTACT)}} & \textbf{98.67} & 75.67 & 84.83 & {\scriptsize\textcolor{black!65}{(Gemini)}} & \textbf{93.30} \\
    Sokoban-Long & 41.50 & 52.00 & {\scriptsize\textcolor{black!65}{(Gemini)}} & \textbf{75.17} & 16.33 & 35.00 & {\scriptsize\textcolor{black!65}{(Gemini)}} & \textbf{52.33} \\
    PushT & 56.00 & 78.00 & {\scriptsize\textcolor{black!65}{(HWM)}} & \textbf{86.17} & 20.00 & 31.00 & {\scriptsize\textcolor{black!65}{(HWM)}} & \textbf{41.33} \\
    Cube-Single & 51.33 & 62.17 & {\scriptsize\textcolor{black!65}{(DINO-WM)}} & \textbf{62.67} & 54.17 & 66.33 & {\scriptsize\textcolor{black!65}{(Fast-LeWM)}} & \textbf{67.83} \\
    \midrule
    Mean & 59.1 & 75.9 & & \textbf{84.4} & 38.6 & 61.4 & & \textbf{69.5} \\
    \bottomrule
  \end{tabular*}
  \par\smallskip
  \parbox{\linewidth}{\footnotesize
  INTACT denotes its Pure CEM variant. Means weight the five tasks equally.}
\end{table}

\noindent\textbf{Strong performance across tasks and action spaces.}\enspace
At offsets $50$ and $100$, Dual-WM averages $84.4\%$ and $69.5\%$,
compared with $75.9\%$ and $61.4\%$ for the task-wise best non-Dual-WM
aggregate without actor-guided proposals. At offset $100$, it exceeds LeWM on every task, improving
the mean by $30.8$ percentage points before rounding.
TwoRoom and Sokoban-Long illustrate the long-range benefit:
Dual-WM reaches $92.50\%$ and $52.33\%$, versus $27.00\%$ and $16.33\%$
for LeWM. Sokoban-Long uses discrete actions in a game-like environment,
extending the evidence beyond continuous control.
At offset $25$, the same model reaches $100\%$, $99.0\%$, and $98.0\%$
on TwoRoom, Reacher, and PushT, respectively.
On PushT, it also attains $86.17\%$ at offset $50$ and $62.00\%$ at
offset $75$, compared with HWM's $78\%$ and $61\%$.
At offset $100$, it reaches $41.33\%$ versus $31.00\%$ for our reproduced HWM,
a gain of $10.33$ percentage points
(Table~\ref{tab:results-pusht}).

\noindent\textbf{Hierarchical planning and complementary action priors.}\enspace
At offset $100$, the full from-scratch configuration exceeds the
low-level-only variant by $23.83$, $4.13$, $6.50$, $21.50$, and $6.33$
percentage points on TwoRoom, Reacher, Sokoban-Long, PushT, and Cube-Single,
respectively; both configurations use the same low-level checkpoint.
With actor-guided proposals on Cube-Single, Dual-WM reaches $82.50\%$
at offset $100$, compared with $80.67\%$ for INTACT (Actor+CEM)
(Table~\ref{tab:actor-comparison}). This comparison uses the INTACT actor
in both methods and is separate from Table~\ref{tab:aggregate}.

\noindent\textbf{Success across planning budgets.}\enspace
On TwoRoom at offset $100$, a four-budget sweep compares success against
measured planning time per environment step (Figure~\ref{fig:planning-budget}).
Dual-WM achieves higher success at lower measured planning time:
$90\%$ at $92.4$\,ms per environment step, versus $84\%$ at
$247.7$\,ms for the low-level-only variant. A larger budget yields
$98\%$ success at $214.0$\,ms.

\subsection{Q2: Do latent goal distances distinguish long-range progress?}
\label{sec:concentration}

We test whether the latent goal distance remains discriminative beyond local
neighborhoods. To compare representations of different widths, we normalize the
Euclidean distance by $\sqrt{2D}$, where $D$ is the corresponding latent
dimension.  The independent isotropic-Gaussian root-mean-square reference is one.
Appendix~\ref{app:concentration-analysis} gives the reference calculation.
Figure~\ref{fig:saturation} shows that LeWM approaches this reference after a
short physical range, whereas the learned low- and high-level spaces preserve
substantially more ordering over long-range TwoRoom paths.  The same pattern is
visible in Reacher configuration space, with the high-level representation
showing the strongest rank correlation.

Spatial diagnostics in Appendix~\ref{app:representation-fields}
support this pattern across navigation, joint configuration, and manipulation.
TwoRoom distance fields preserve goal-directed variation across rooms
(Figure~\ref{fig:heatmap}). On Reacher, high-level goal distances correlate
more strongly with wrapped joint-angle distance than the low-level and
LeWM representations (Spearman $0.562$, $0.318$, and $0.190$).
On PushT, Dual-WM-H gives the largest improvements in position correlation
and both pairwise goal-ordering measures, although its orientation
correlation is statistically indistinguishable from LeWM.
The benefit is therefore clearest in goal discrimination, rather than
uniform improvement in every physical-distance correlation.

\subsection{Q3: Does rollout training preserve physical-state information?}
\label{sec:physical-state}
To test whether recursively predicted states retain task information,
we compare one-step and rollout-trained low-level checkpoints with
horizon-resolved physical probes. Separate ridge probes are fitted on
predicted latents for each checkpoint and target, with disjoint episodes
for fitting, validation, and evaluation
(Appendix~\ref{app:representation-fields}).
The rollout-trained checkpoint has lower point estimates at every
horizon for all four quantities (Figure~\ref{fig:rollout-physical-probe}).
The separation is clearest for Reacher joint configuration and fingertip
position, and PushT block position; PushT orientation has wider
uncertainty. These results support the role of recursive training in
retaining task-relevant state information during multi-step prediction.

\subsection{Q4: Which design choices improve planning success?}
\label{sec:ablations}

On TwoRoom, we test the latent-space design and LoRe scale activation
with the full hierarchical planner. Horizon and weighting studies use
low-level-only planning to isolate low-level rollout training; MAPS is
evaluated with the hierarchical planner. Planning results average three
planning seeds and are shown without error bars. Within each study,
unablated settings remain at their defaults.

\begin{figure}[t]
  \begin{minipage}[t]{0.48\linewidth}
    \centering
    \includegraphics[width=\linewidth]{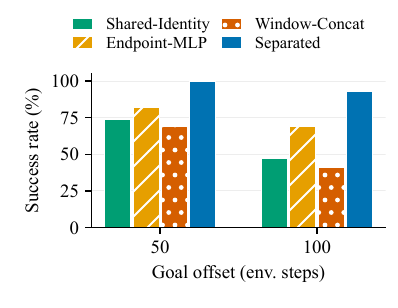}
    \caption{Latent-space design on TwoRoom. All variants share the low-level checkpoint and hierarchical planning. Window-Concat directly concatenates the same window used by Separated (Dual-WM). Bars show mean success.}
    \label{fig:ablation-latent-space}
  \end{minipage}\hfill
  \begin{minipage}[t]{0.48\linewidth}
    \centering
    \includegraphics[width=\linewidth]{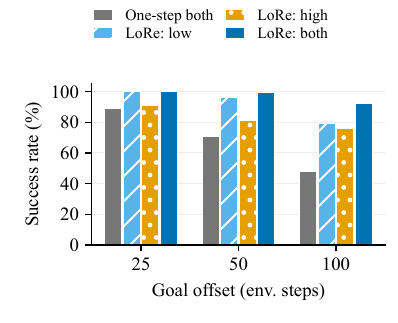}
    \caption{LoRe scales on TwoRoom. Both model levels and hierarchical planning are retained; inactive levels use one-step supervision. Bars show mean success.}
    \label{fig:ablation-lore-scales}
  \end{minipage}
\end{figure}

\noindent\textbf{Learned window interfaces improve planning.}\enspace
Shared-Identity and Endpoint-MLP use endpoint latents; Window-Concat
directly concatenates the same length-$k$ window used by Dual-WM.
All retain the same low-level checkpoint and coarse-to-fine planner.
At offset $100$, Dual-WM achieves $93\%$ success versus $41\%$ for
Window-Concat, a $52$-percentage-point gap with the same input window.
Endpoint-MLP and Shared-Identity achieve $69\%$ and $47\%$, respectively
(Figure~\ref{fig:ablation-latent-space}). In the farthest distance bin,
Dual-WM-H reaches $84.9\%$ goal-ordering accuracy versus $51.0\%$ for
Window-Concat (Figure~\ref{fig:ablation-goal-ordering},
Appendix~\ref{app:ablation-protocol}). Together, these results support
learning a high-level representation for long-range goal evaluation,
rather than relying on window context alone.

\noindent\textbf{LoRe benefits both temporal scales.}\enspace
We retain both dynamics models and hierarchical planning, varying only
which levels receive LoRe instead of one-step supervision
(Figure~\ref{fig:ablation-lore-scales}). At offset $100$, dual-scale LoRe
raises success from $48\%$ with one-step supervision to $92\%$.
Low- and high-level LoRe alone yield $79\%$ and $76\%$,
respectively.
Both-scale training improves over the stronger single-scale setting by
$13$ percentage points. Low-level LoRe already reaches $100\%$ at offset
$25$, whereas the benefit of also supervising high-level rollouts is
clearest at longer offsets. These are training switches, not removal
of either planning branch.

\noindent\textbf{Low-level rollout length and weighting.}\enspace
Under low-level-only planning, offset-$100$ success rises from $32\%$
at $N=1$ to $69\%$ at $N=5$, then falls to $37\%$ at $N=10$.
At $N=5$, exponential weighting gives $69\%$, versus $62\%$ for linear
and $54\%$ for uniform weights. These studies favor moderate rollout
lengths with decayed supervision. Full sweeps and physical-state probes
are in Appendix~\ref{app:ablation-protocol}.

\noindent\textbf{Macro-action prior shaping.}\enspace
At offset $100$, deterministic macro-action encoding achieves $67\%$
success, versus $81\%$ for stochastic encoding without MAPS ($\beta=0$).
Adding MAPS raises success to $93.5\%$ at $\beta=10^{-2}$, despite higher
one-step and six-step latent MSE than at $\beta=0$. This pattern is
consistent with MAPS shaping a macro-action space better suited to search,
rather than simply minimizing prediction error. Further increasing
$\beta$ to $5\times10^{-2}$ reduces success to $91.5\%$
(Figure~\ref{fig:ablation-maps}, Appendix~\ref{app:ablation-protocol}).

\section{Conclusion}
\label{sec:conclusion}

Long-horizon latent planning requires both reliable recursive predictions
and goal distances that distinguish progress beyond local neighborhoods.
\dlwm{} addresses these requirements through distinct representations and
dynamics for local execution and long-range planning, coupled by a learned
window interface. LoRe trains recursive predictions at both scales, while
MAPS organizes the macro-action space for search. Across five visual
control tasks spanning continuous and discrete actions, this design
improves extended-offset goal reaching. The representation diagnostics
and ablations support complementary roles for long-range goal discrimination,
recursive consistency, and precise low-level convergence. Together, these
findings suggest that the usefulness of a world model for planning depends
on how its representations, predictive training, and goal-evaluation metrics are
organized across temporal scales.

Our evaluation uses offline data and fixed temporal abstraction, and gains
depend on the task and action-proposal mechanism. Adaptive temporal scales
and evaluation beyond the present offline setting are natural next steps.

\clearpage
\setlength{\parskip}{6pt}

\bibliography{references}
\bibliographystyle{dualwm}

\FloatBarrier
\clearpage
\appendix

\raggedbottom
\setlength{\floatsep}{10pt plus 2pt minus 2pt}
\setlength{\textfloatsep}{12pt plus 2pt minus 2pt}
\setlength{\intextsep}{10pt plus 2pt minus 2pt}
\makeatletter
\setlength{\@fptop}{0pt}
\setlength{\@fpsep}{14pt plus 2pt minus 2pt}
\setlength{\@fpbot}{0pt plus 1fil}
\makeatother

\section{Latent Distance Concentration: Analysis}
\label{app:concentration-analysis}

Consider \(z\in\mathbb{R}^d\) with independent components distributed as
\(\mathcal{N}(0,1)\). Its squared norm follows a chi-squared distribution,
\[
  \|z\|^2=\sum_{i=1}^{d}z_i^2\sim\chi^2_d,\qquad
  f(x;d)=\frac{x^{d/2-1}e^{-x/2}}{2^{d/2}\Gamma(d/2)},\quad x>0.
\]
Consequently,
\[
  \mathbb{E}\!\left[\|z\|^2\right]=d,\qquad
  \mathrm{Var}\!\left(\|z\|^2\right)=2d,\qquad
  \frac{\sqrt{\mathrm{Var}(\|z\|^2)}}{\mathbb{E}[\|z\|^2]}
  =\sqrt{\frac{2}{d}}.
\]
Thus the norm has approximate standard deviation \(1/\sqrt{2}\) and relative
fluctuation \(1/\sqrt{2d}\); for \(d=192\), it concentrates near
\(\sqrt{192}\approx13.86\).

For independent \(x,y\in\mathbb{R}^d\) with the same standard-normal
components, \(x-y\) has components distributed as \(\mathcal{N}(0,2)\), so
\[
  \|x-y\|^2\sim 2\chi^2_d,\qquad
  \mathbb{E}\!\left[\|x-y\|^2\right]=2d,\qquad
  \mathrm{Var}\!\left(\|x-y\|^2\right)=8d.
\]
The distance itself has
\[
  \mathbb{E}\!\left[\|x-y\|\right]
  =2\frac{\Gamma((d+1)/2)}{\Gamma(d/2)},\qquad
  \mathrm{Var}(\|x-y\|)=2d-\mathbb{E}[\|x-y\|]^2.
\]
For \(d=192\), the exact mean is approximately \(19.5704\), close to
\(\sqrt{2d}=\sqrt{384}\approx19.5959\), and the standard deviation is about
\(0.9993\). These values provide an independent-Gaussian reference for
evaluating distance contrast in learned representations.

A useful distinction is between marginal regularization and temporal
dependence. If encoded states have zero mean and identity covariance, define
\(\rho_h=d^{-1}\mathbb{E}[z_t^\top z_{t+h}]\). Then
\begin{equation}
  \mathbb{E}\|z_t-z_{t+h}\|_2^2=2d(1-\rho_h).
  \label{eq:correlation-distance}
\end{equation}
As \(\rho_h\) decreases toward zero, the mean squared distance approaches
\(2d\). Gaussian marginals alone do not imply independence, and one-step
training does not mathematically force temporal correlations to vanish.
Rather, it supplies no explicit requirement that remote-state distances rank
task progress; a plateau near the reference is consistent with weakened
temporal discrimination in the planner's distance.

When norms concentrate, cosine similarity and squared Euclidean distance carry
closely related information:
\[
  \|x-y\|_2^2=\|x\|_2^2+\|y\|_2^2-2\|x\|_2\|y\|_2\cos(x,y).
\]
Changing between these metrics therefore need not restore useful contrast.
Learned costs or additional supervision may exploit task information that
the original metric underweights. The analysis therefore concerns the
contrast available to the planner's chosen distance.

\subsection{Candidate ranking under prediction error}
\label{app:ranking-margin}

Fix a latent space, a goal $z_g$, and two candidate control sequences
$B_1,B_2$ over the same execution horizon. Let $z_i$ be the encoded endpoint
of the actual trajectory under $B_i$, and $\hat z_i$ the corresponding model
prediction from the same initial observation. We condition on the realized
trajectories if transitions are stochastic. Define
\begin{equation}
  c_i=\|z_i-z_g\|_2,\qquad
  \hat c_i=\|\hat z_i-z_g\|_2,\qquad
  \|\hat z_i-z_i\|_2\leq\varepsilon_i,\quad i\in\{1,2\}.
  \label{eq:ranking-cost-error}
\end{equation}
The unsquared distances simplify the analysis and induce the same candidate
ordering as the squared distances used by the planner. The reverse triangle
inequality gives $|\hat c_i-c_i|\leq\varepsilon_i$.

\paragraph{Pairwise ranking condition.}
Suppose $c_1<c_2$, with margin $m=c_2-c_1$. Then
\begin{equation}
  m>\varepsilon_1+\varepsilon_2
  \quad\Longrightarrow\quad
  \hat c_1<\hat c_2.
  \label{eq:ranking-margin}
\end{equation}
Indeed,
\[
  \hat c_2-\hat c_1
  \geq(c_2-\varepsilon_2)-(c_1+\varepsilon_1)
  =m-\varepsilon_1-\varepsilon_2>0.
\]
This sufficient condition connects the two requirements for long-horizon
planning. Recursive prediction error increases the uncertainty in candidate
costs, while reduced goal-distance contrast can shrink the margin available
to distinguish them. A smaller margin therefore requires a tighter prediction
error bound to certify the same ordering. The condition is sufficient;
smaller or cancelling cost errors can also preserve the ranking.

For primitive-action candidates at horizon $h$, the low-level bound in
Eq.~(\ref{eq:error-propagation}) gives, under its assumptions, the common
endpoint-error bound and sufficient margin
\[
  \varepsilon_1=\varepsilon_2=\epsilon_L\sum_{j=0}^{h-1}L_L^j,
  \qquad
  m>2\epsilon_L\sum_{j=0}^{h-1}L_L^j
\]
when both candidates satisfy the same
residual and sensitivity bounds. This relates prediction accuracy to the
resolution required by the goal cost.

The condition applies to candidates with defined actual endpoints and
compares their ordering under the chosen latent cost. Its connection to task
progress is examined through the goal-geometry diagnostics in
Section~\ref{sec:concentration}. For nonzero total error, the ratio
$m/(\varepsilon_1+\varepsilon_2)$ is invariant to a common positive
rescaling of the latent space.

\section{Finite-Horizon Rollout Weighting}
\label{app:rollout-weighting}

This section develops the weighting rationale in Section~\ref{sec:lore}
by analyzing error propagation for fixed predictors over finite rollout
horizons and the resulting trade-off in supervision across horizons.

\subsection{Propagation along a fixed trajectory}
Suppress the scale index and fix the controls, including any sampled
macro-actions. Let $x_h$ denote the encoded input state after $h$ steps,
and let $F_h$ be the learned update with the corresponding control held
fixed. For a single-state predictor, $x_h=z_{t+h\Delta}$ and $F_h$ is $P$
with that control supplied. For a history-conditioned predictor, $x_h$
stacks the latent history, and $F_h$ shifts this history and appends the
predicted latent. Any observed initial context and control history are
shared by the encoded and predicted sequences.

Write $\hat x_{h+1}=F_h(\hat x_h)$, $\hat x_0=x_0$, and suppose
\begin{equation}
  \|F_h(\hat x_h)-F_h(x_h)\|_2
       \leq L\|\hat x_h-x_h\|_2,\qquad
  r_{h+1}:=\|F_h(x_h)-x_{h+1}\|_2\leq\epsilon
  \label{eq:rollout-assumptions}
\end{equation}
for the steps considered, with $L\geq0$.
The first condition must hold across the compared encoded and predicted
inputs, not only on data states. The second measures a residual along the
realized trajectory; it does not require an exact Markov dynamics map in
the compressed latent space. For history inputs, $L$ bounds the complete
shift-and-predict update, not just the predictor's dependence on its newest
input.

\paragraph{Finite-horizon bound.}
Let $d_h=\|\hat x_h-x_h\|_2$. The triangle inequality gives
$d_{h+1}\leq Ld_h+r_{h+1}$. Induction from $d_0=0$ yields
\begin{equation}
  d_h\leq\sum_{i=1}^h L^{h-i}r_i
      \leq\epsilon S_h(L),\qquad
  S_h(L)=
  \begin{cases}
    (L^h-1)/(L-1), & L\ne1,\\
    h, & L=1.
  \end{cases}
  \label{eq:finite-rollout-bound}
\end{equation}
The newest latent is a coordinate block of $x_h$, so its Euclidean error
is at most $d_h$. This recovers the endpoint bound in
Eq.~(\ref{eq:error-propagation}) for either input convention.
The constants are conditional bounds on the rollouts considered, not
measured global properties of our models. No cancellation of the local
residuals is assumed, so the bound can be conservative.

\subsection{A finite-horizon reference for weighting}
Let $n$ denote the rollout length of the scale under consideration,
with $n=N$ at the low level and $n=M$ at the high level.
For fixed $n$, consider equalizing contributions of the squared upper
envelope $[\epsilon S_h(L)]^2$. Since $S_h(L)>0$ for $h\geq1$,
the following normalized weights equalize these contributions at every horizon:
\begin{equation}
  v_h=\frac{S_h(L)^{-2}}{\sum_{j=1}^n S_j(L)^{-2}}
  \quad\text{satisfy}\quad
  v_h[\epsilon S_h(L)]^2
    =\frac{\epsilon^2}{\sum_{j=1}^n S_j(L)^{-2}}
  \label{eq:envelope-reference}
\end{equation}
This identity balances the squared-error upper envelope at fixed model
parameters. We use its asymptotic behavior to motivate the exponential
weighting family implemented in Eq.~(\ref{eq:lore}).

For $L>1$ and $L^h\gg1$, $S_h(L)^{-2}\sim(L-1)^2L^{-2h}$.
The reference's adjacent-horizon decay is
\begin{equation}
  -\log\frac{v_{h+1}}{v_h}
  =2\log\frac{S_{h+1}(L)}{S_h(L)},\qquad 1\leq h<n,
  \label{eq:effective-envelope-decay}
\end{equation}
which approaches $2\log L$ in the expanding regime.
At $L=1$, $v_h\propto h^{-2}$; for $0\leq L<1$, the unnormalized
reference $S_h(L)^{-2}$ tends to $(1-L)^2$.
When $h|L-1|\ll1$, $S_h(L)\approx h$ even if $L>1$.
Finite training horizons therefore need not reach the exponential regime,
which is another reason not to prescribe $\alpha=2\log L$.

\subsection{Decay and the balance of horizon supervision}
Even if $L$ and $\epsilon$ were known, reducing a weighted upper bound
alone would not identify a useful balance of supervision. With these
constants and $n$ fixed, define
\begin{equation}
  U_n(\alpha,L)=\epsilon^2\sum_{h=1}^n w_h(\alpha)S_h(L)^2,
  \qquad
  \frac{\partial U_n}{\partial\alpha}
    =-\epsilon^2\operatorname{Cov}_{w(\alpha)}
        \bigl(h,S_h(L)^2\bigr)\leq0.
  \label{eq:envelope-decay-tradeoff}
\end{equation}
The derivative follows from
$\partial_\alpha w_h=w_h(\sum_{j=1}^n j w_j-h)$.
The covariance is nonnegative because both arguments are nondecreasing
in $h$; explicitly it equals
$\frac12\sum_{i,j}w_iw_j(i-j)[S_i(L)^2-S_j(L)^2]\geq0$.
Thus larger decay never increases this envelope objective, even though
it suppresses the compositions that multi-step training is meant to constrain.

To quantify the effect on terminal supervision, fix the
model and let $E_h$ be its expected squared rollout error under the same
data and macro-action distribution as Eq.~(\ref{eq:lore}). Assuming these
expectations are finite, nonnegativity and the limit of the weights give
\begin{equation}
  E_n\leq\frac{\mathcal L_{\mathrm{LoRe}}}{w_n},\qquad
  \lim_{\alpha\to\infty}\mathcal L_{\mathrm{LoRe}}=E_1
  \quad\text{for fixed }n>1.
  \label{eq:terminal-supervision}
\end{equation}
As $w_n$ becomes small, a given bound on the weighted loss gives a weaker
bound on terminal error, and the limiting objective is one-step prediction.
Both this relation and the derivative in
Eq.~(\ref{eq:envelope-decay-tradeoff}) hold at fixed model parameters;
the effect of retraining is assessed empirically.

\paragraph{Time units in the two scales.}
For a horizon of $\tau=hf\Delta_s$ environment steps, the unnormalized weight is
$e^{-\alpha_s h}=e^{-(\alpha_s/(f\Delta_s))\tau}$.
A common relative decay per environment step, $\eta$, would imply
$\alpha_L=f\eta$ and $\alpha_H=kf\eta$, hence $\alpha_H=k\alpha_L$.
This only matches physical
decay rates; the sampled horizons and normalization constants can still
differ. Moreover, the two predictors can have different residuals and
sensitivities. We therefore parameterize their decays separately.

\section{Extended Related Work}
\label{app:related}

\subsection{Latent world models for control}
Latent dynamics support planning from pixels and policy learning in imagined
trajectories \citep{hafner2019planet,hafner2023dreamerv3,hansen2023tdmpc2}.
Our setting is closest to reconstruction-free, goal-conditioned planning:
DINO-WM predicts pretrained visual features \citep{zhou2024dinowm}, PLDM learns
dynamics from reward-free offline trajectories \citep{sobal2025pldm}, and
LeWM learns an encoder and predictor jointly with an anti-collapse regularizer
\citep{maes2026leworldmodel}. These methods establish the predictive latent
interface on which we build. \dlwm{} changes how that interface is organized
across temporal scales: the representation used for primitive transitions need
not also define the geometry used to evaluate distant subgoals.

\subsection{Prediction beyond one-step training}
Long-horizon prediction can be improved by changing the prediction interface or
by training on recursively generated states. VLWM predicts outcomes of
variable-length action sequences \citep{du2026vlwm}; Fast-LeWM predicts
action-prefix outcomes in parallel from an observed anchor
\citep{gao2026fastlewm}; Flow-JEPA generates future latent sequences with
conditional flow matching \citep{huo2026flowjepa}. These approaches reduce
dependence on a long chain of local transitions. Open-loop training instead
retains recursive dynamics and supervises the resulting trajectory. HWM's
stated autoregressive objective sums rollout errors uniformly across horizons
\citep{zhang2026hwm}. RC-aux formulates general horizon-weighted open-loop
prediction \citep{li2026rcaux}; its released default implementation uses
normalized linear weights that increase with the horizon.
In contrast, \lore{} uses geometrically decaying weights motivated by the
error-amplification envelope of recursive prediction. It applies separate
decay profiles to primitive-action and macro-action dynamics, so both are
trained on self-generated trajectories while balancing the influence of
near- and distant-horizon errors
(Section~\ref{sec:lore}).

\subsection{Representation geometry for goal evaluation}
Accurate prediction and state decodability do not guarantee that a latent
distance ranks plans correctly. DA-LeWM studies decision-metric alignment and
uses action-conditioned auxiliary objectives \citep{wang2026dalewm}; SCALE
aligns latent distances with distances in task-relevant state variables
\citep{hu2026scale}. RC-aux learns budget-conditioned reachability
\citep{li2026rcaux}, while TD-JEPA mines a directed temporal cost from trajectory
order \citep{bai2026tdjepa}. These objectives can reshape representations as well
as change planning costs. We instead use a dedicated high-level latent state space
with predictive supervision across extended transitions, without introducing
a state-distance target or a learned reachability head.

SD-JEPA is especially relevant: it separates progression and content
coordinates within one embedding and supplies explicit temporal triplet
supervision \citep{thil2026sdjepa}. \dlwm{} separates \emph{dynamical roles}:
each state space has its own predictor and action timescale, and the high-level
encoder also maps low-level predictions into the subgoal space during
planning. The resulting separation governs both prediction and goal matching:
long-range subgoals are evaluated in the high-level latent state space, while final
goal convergence uses the low-level latent state space. Section~\ref{sec:concentration}
reports the empirical geometry diagnostic; Appendix~\ref{app:concentration-analysis}
provides its concentration reference.

\subsection{Temporal abstraction and hierarchical planning}
HWM learns dynamics at multiple temporal resolutions within a shared latent
space and transfers predicted subgoals directly to low-level MPC
\citep{zhang2026hwm}. Hi-LeWM adds macro-action dynamics on top of a frozen
LeWM encoder and low-level predictor while retaining that latent interface
\citep{caselli2026hilewm}. Both are close precedents for our macro-action
planner. \dlwm{} changes the coupling: a learned map $E_H$ produces a distinct
high-level state, and primitive-action plans are evaluated by mapping their
predicted latent windows into that space. The final approach to the goal
uses the low-level distance directly. This permits coarse subgoal matching and
fine final control to use different geometries while retaining a common visual
input. Table~\ref{tab:positioning} situates these hierarchical approaches
relative to the flat LeWM baseline; additional work on prediction, search, and
model reliability is discussed in Appendix~\ref{app:related}.

\begin{table}[t]
\centering
\caption{LeWM and hierarchical latent world models with learned macro-actions.}
\label{tab:positioning}
\small
\setlength{\tabcolsep}{5pt}
\renewcommand{\arraystretch}{1.05}
\begin{tabular}{@{}llll@{}}
\toprule
Method & State spaces & Dynamics trained & Prediction training \\
\midrule
LeWM & Single & Primitive & One-step transition \\
HWM & Shared across scales & Primitive + macro & Uniform-weight rollout \\
Hi-LeWM & Shared; low frozen & Macro only & One-step waypoint \\
\textbf{Dual-WM} & \textbf{Separate across scales} & \textbf{Primitive + macro} & \textbf{Scale-specific decayed rollout} \\
\bottomrule
\end{tabular}
\par\smallskip
\end{table}

\subsection{Amortizing test-time search}
A complementary line of work reduces the cost of test-time action selection. \citet{sun2026intact} train an end-to-end intent-to-action
interface that produces actions without test-time search, with an optional local
CEM that uses far fewer sampled sequences than a global search.
\citet{wang2026prism} attach a state-conditioned Gaussian action prior to a frozen
encoder and fuse it into the planner's sampling distribution in closed form, so
that search concentrates where the prior is confident, and \citet{gao2026imwm}
pair the world model with a retrieval-and-scoring intuition model, showing that
even an idealized predictor still fails under a finite search budget.
\citet{rakhimov2026qantara}
and \citet{sommer2026rp1} learn the search itself, the latter training a critic and
an optimizer entirely on imagined rollouts of a pretrained model.
\citet{huang2026leflow} amortize planning into a reusable latent trajectory prior
that a flow model proposes and a frozen world model verifies, and
\citet{nguyen2026latentgeometry} show that with a suitably regularized geometry the
planner can be replaced outright by a goal-conditioned inverse dynamics model.
\citet{pham2026leap} instead keep CEM in the loop but optimize the whole action
horizon as a differentiable variable through a frozen model. These methods are
complementary to ours: they change how candidate actions are proposed or refined,
whereas we change the latent space and the rollout supervision in which candidates
are evaluated.

\subsection{Reliability and diagnosis}
A separate line asks when latent planning fails rather than how to improve it, and
provides the diagnostics that motivate our analysis. \citet{wu2026viscore}
decompose planning-relevant quality into the reachability and capacity of the
predictor given encoded features plus planner hallucination, and find these
correlate with success far better than latent straightness or probing accuracy.
\citet{an2026acpc} use bisimulation to measure how far a clean history and a
visually perturbed view diverge under a shared action sequence, turning
perturbation robustness into a screening tool. \citet{vakalis2026intervention}
show that reward fit neither reveals nor guarantees intervention fidelity, and
\citet{wang2026adajepa} adapt a world model online from the transitions it
actually observes when plans fail. In contact-rich settings,
\citet{zheng2026contactguard} roll a policy's planned chunk forward in latent space
to abort before contact rather than after. We adopt the diagnostic stance of this
line: our analysis of distance saturation and of per-horizon rollout error is
intended as a measurement of the two failure modes, not only as a motivation for
the architecture.

\subsection{Theory}
Several recent results bound what a latent predictive model can and cannot
guarantee. \citet{klindt2026when} prove linear identifiability of the true latent
degrees of freedom for a class of stationary additive-noise worlds, and show that
the Gaussian is the unique latent distribution for which this holds, which gives a
principled reading of the isotropic-Gaussian regularizer our model inherits.
\citet{cui2026generalization} link JEPA pretraining error to downstream planning
regret through an action-conditioned co-occurrence factorization, exposing a
latent-dimension trade-off between approximation and sample error.
\citet{you2026controltheory} recast model selection as the gap between predicted
and true plan cost at the committed plan, and \citet{arnez2026sigreg} place the
isotropic Gaussian state-regularization objective in an active-inference
hierarchy of anti-collapse regularizers.
Our analysis in Appendix~\ref{app:concentration-analysis} is of the same
diagnostic character: it characterizes a reference regime in which distances
can lose contrast for ordering candidates. The complementary analysis in
Section~\ref{sec:lore} connects recursive error propagation to horizon weighting.

\section{Training and Planning Configurations}
\label{app:configuration}

\paragraph{Data and training.}
Table~\ref{tab:training-config} summarizes the from-scratch configuration.
Each dataset is split $9{:}1$ into training and test episodes.
Images are resized to $224\times224$ and use ImageNet normalization,
without cropping or data augmentation. We use AdamW with weight decay
$10^{-3}$, prediction-loss weights $\lambda_L=\lambda_H=1$, and MAPS
coefficient $\beta=10^{-2}$ on all tasks. The high-level state-regularization weight
is $0.30$; the low-level weights are listed in the table.

\begin{table}[!htbp]
  \centering
  \caption{Task-specific training configurations. Paired entries are low/high
  level. Environment-step counts are approximate dataset totals; $N,M$
  count model transitions at their respective scales.}
  \label{tab:training-config}
  \scriptsize
  \setlength{\tabcolsep}{4pt}
  \renewcommand{\arraystretch}{1.16}
  \begin{tabular}{@{}lccccc@{}}
    \toprule
    Setting & TwoRoom & Reacher & Sokoban-Long & PushT & Cube-Single \\
    \midrule
    Episodes & 10,000 & 10,000 & 2,500 & 18,685 & 10,000 \\
    Environment steps & 921,000 & 2,010,000 & 395,000 & 2,337,000 & 2,010,000 \\
    $f$ & 5 & 5 & 1 & 5 & 5 \\
    $k$ & 3 & 3 & 2 & 2 & 2 \\
    $N/M$ & 5/6 & 3/6 & 10/8 & 3/6 & 6/6 \\
    $\alpha_L/\alpha_H$ & 0.5/0.5 & 0.2/0.5 & 0.1/0.5 & 0.2/0.5 & 0.05/0.5 \\
    Low-level regularization weight & 0.30 & 0.30 & 0.25 & 0.09 & 0.15 \\
    Low-level learning rate & $5\times10^{-5}$ & $2\times10^{-5}$ & $2.5\times10^{-5}$ & $10^{-5}$ & $5\times10^{-6}$ \\
    High-level learning rate & $5\times10^{-5}$ & $5\times10^{-5}$ & $5\times10^{-5}$ & $3\times10^{-5}$ & $5\times10^{-5}$ \\
    Batch size (L/H) & 256/256 & 256/256 & 512/512 & 256/512 & 512/256 \\
    Epochs (L/H) & 5/5 & 2/5 & 5/5 & 6/6 & 5/5 \\
    \bottomrule
  \end{tabular}
\end{table}

\paragraph{Model architecture.}
The from-scratch visual encoder $E_L$ is a 12-layer ViT-Tiny with
$14\times14$ patches and width $192$, producing $d_L=192$ latent states.
The low-level predictor $P_L$ is a six-layer, 16-head causal Transformer
with latent width $192$ and a three-state history; actions condition the
predictor through adaptive layer normalization.
The high-level encoder $E_H$ is a four-layer, eight-head bidirectional
Transformer over $k$ low-level states, using the final token as its
$d_H=192$ output. The high-level predictor $P_H$ also has six layers,
16 heads, and latent width $192$, with context consisting of the current
high-level state and macro-action.
The macro-action encoder $E_A$ uses a four-layer, eight-head bidirectional
Transformer and mean pooling to parameterize a diagonal-Gaussian posterior
with $d_u=16$. Its sampled macro-action is projected to width $192$ for
conditioning $P_H$.

\paragraph{VFM-adaptor variant.}
This variant freezes DINOv2-S/14 and learns a width-$384$ adaptor.
Pixel patch tokens attend to frozen DINO tokens through two six-head
cross-attention layers, with zero-initialized feature modulation and
MLP blocks of expansion ratio four. Attention pooling with one learned
query and an MLP projection head with batch normalization produce the
low-level state. The adaptor and low-level predictor are trained with
LoRe, supplemented by spatial-structure and semantic-consistency losses
weighted $0.1$ each. High-level training then follows the from-scratch
procedure. Task-specific training hyperparameters follow
Table~\ref{tab:training-config}.

\paragraph{Planning.}
Table~\ref{tab:planning-config} lists the planning horizons. On all tasks,
low-level CEM uses $300$ candidates per iteration, $30$ iterations, and
$30$ elites; high-level CEM uses $100$ candidates, $20$ iterations, and
$10$ elites. Sokoban-Long uses categorical low-level search, while the
other tasks use Gaussian low-level search; high-level macro-action
search is Gaussian throughout. At goal offsets $25$, $50$, $75$, and
$100$, the maximum execution budgets are $50$, $75$, $100$, and $125$
environment steps, respectively, wherever the offset is evaluated.

\begin{table}[!htbp]
  \centering
  \caption{Planning horizons. $H_L$ counts low-level model steps and
  $H_H$ counts high-level macro-steps.}
  \label{tab:planning-config}
  \small
  \setlength{\tabcolsep}{7pt}
  \begin{tabular}{@{}lccccc@{}}
    \toprule
    & TwoRoom & Reacher & Sokoban-Long & PushT & Cube-Single \\
    \midrule
    $H_L$ & 8 & 5 & 12 & 5 & 5 \\
    $H_H$ & 6 & 6 & 8 & 6 & 6 \\
    \bottomrule
  \end{tabular}
\end{table}

\FloatBarrier

\section{Full planning results}
\label{app:full-results}

\begin{figure}[t]
  \centering
  \includegraphics[width=0.185\linewidth]{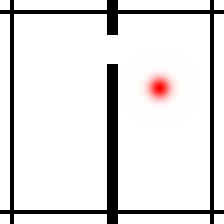}\hfill
  \includegraphics[width=0.185\linewidth]{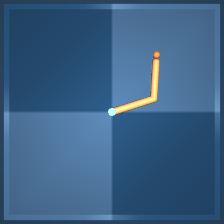}\hfill
  \includegraphics[width=0.185\linewidth]{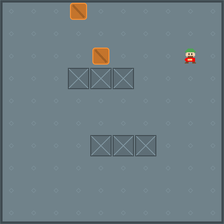}\hfill
  \includegraphics[width=0.185\linewidth]{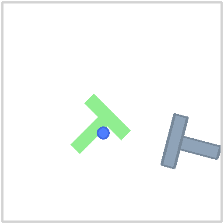}\hfill
  \includegraphics[width=0.185\linewidth]{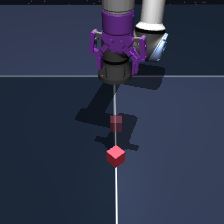}
  \caption{The evaluation environments: TwoRoom, Reacher, Sokoban-Long, PushT
  and Cube-Single (left to right).}
  \label{fig:envs}
\end{figure}

Tables~\ref{tab:results-tworoom}--\ref{tab:results-cube} report all
task-level results and goal offsets. Entries marked $\dagger$ are our
reproduced evaluations; quoted entries retain their original reporting
conventions. Our evaluations use three planning seeds, with means and
standard deviations computed across seed-specific task sets and search
randomness. Within each seed, methods share the same evaluation tasks and
environment-step budgets (Appendix~\ref{app:baselines}). Bold marks a
Dual-WM variant only when it attains the highest reported mean in the
column, including ties. A dash denotes an
unavailable result.

\paragraph{Dual-WM variants.}
From scratch jointly trains the visual encoder and low-level predictor,
with the high-level branch learning from their representations.
VFM adaptor freezes a pretrained visual encoder and trains an adaptor
together with the low-level predictor, followed by high-level training.
Low-level only plans with primitive
actions without high-level subgoal guidance, using the same low-level
checkpoint as the full model. With INTACT actor adds
actor-guided proposals to low-level search on Cube-Single. The main
cross-task comparison always uses the from-scratch variant.

\begin{table}[t]
\centering
\caption{Goal-reaching success rates (\%) on TwoRoom (left) and
Reacher (right); columns are goal offsets measured in original environment steps.
$^{\dagger}$ denotes results reproduced by us.}
\label{tab:results-tworoom}\label{tab:results-reacher}
\scriptsize
\setlength{\tabcolsep}{2.5pt}
\renewcommand{\arraystretch}{1.05}
\begin{tabular}{@{}lcccccc@{}}
  \toprule
  \multirow{2}{*}{Method} & \multicolumn{3}{c}{TwoRoom} & \multicolumn{3}{c}{Reacher} \\
  \cmidrule(lr){2-4} \cmidrule(lr){5-7}
  & 25 & 50 & 100 & 25 & 50 & 100 \\
  \midrule
  LeWM & 87.0 & $61.50\pm1.22$\textsuperscript{$\dagger$} & $27.00\pm0.82$\textsuperscript{$\dagger$} & 86.0 & $85.33\pm0.62$\textsuperscript{$\dagger$} & $75.67\pm0.62$\textsuperscript{$\dagger$} \\
  DINO-WM & 100.0 & $60.50\pm0.71$\textsuperscript{$\dagger$} & $32.17\pm0.94$\textsuperscript{$\dagger$} & 79.0 & $67.00\pm0.71$\textsuperscript{$\dagger$} & $34.17\pm1.03$\textsuperscript{$\dagger$} \\
  HWM & 100\textsuperscript{$\dagger$} & $73.67\pm1.03$\textsuperscript{$\dagger$} & $48.33\pm0.94$\textsuperscript{$\dagger$} & -- & -- & -- \\
  PLDM & 97 & -- & -- & 78 & -- & -- \\
  Fast-LeWM & 98 & $67.17\pm1.25$\textsuperscript{$\dagger$} & $38.67\pm0.62$\textsuperscript{$\dagger$} & 90.0 & $81.33\pm0.94$\textsuperscript{$\dagger$} & $79.33\pm0.62$\textsuperscript{$\dagger$} \\
  RC-aux & $98.0\pm1.4$ & $66.33\pm1.18$\textsuperscript{$\dagger$} & $21.50\pm0.82$\textsuperscript{$\dagger$} & $87.2\pm6.4$ & $85.50\pm1.41$\textsuperscript{$\dagger$} & $81.67\pm1.03$\textsuperscript{$\dagger$} \\
  INTACT\textsuperscript{\textit{Pure CEM}} & $82.89\pm0.84$ & $24.00\pm1.08$\textsuperscript{$\dagger$} & $3.50\pm0.71$\textsuperscript{$\dagger$} & $83.67\pm0.67$ & $92.33\pm1.03$\textsuperscript{$\dagger$} & $81.17\pm0.85$\textsuperscript{$\dagger$} \\
  INTACT\textsuperscript{\textit{Actor+CEM}} & $98.00\pm1.15$ & $73.33\pm0.94$\textsuperscript{$\dagger$} & $63.00\pm1.08$\textsuperscript{$\dagger$} & $86.67\pm0.88$ & $97.33\pm0.62$\textsuperscript{$\dagger$} & $93.00\pm0.82$\textsuperscript{$\dagger$} \\
  VLWM & 96 & 80 & 68 & -- & -- & -- \\
  Gemini controller & $98.00\pm0.41$\textsuperscript{$\dagger$} & $94.83\pm0.85$\textsuperscript{$\dagger$} & $90.00\pm0.82$\textsuperscript{$\dagger$} & $89.00\pm0.41$\textsuperscript{$\dagger$} & $86.17\pm0.47$\textsuperscript{$\dagger$} & $84.83\pm0.62$\textsuperscript{$\dagger$} \\
  \midrule
  \multicolumn{7}{@{}l}{\textbf{Dual-WM}} \\
  \hspace{0.5em}From scratch & \textbf{100.0} & $99.33\pm0.94$ & $92.5\pm0.71$ & $99.0\pm0.82$ & $98.67\pm0.62$ & $93.3\pm1.03$ \\
  \hspace{0.5em}VFM adaptor & \textbf{100.0} & \textbf{\boldmath$99.67\pm0.47$} & \textbf{\boldmath$93.33\pm0.24$} & \textbf{\boldmath$99.33\pm0.24$} & \textbf{\boldmath$98.83\pm0.24$} & \textbf{\boldmath$98.5\pm0.4$} \\
  \hspace{0.5em}Low-level only & \textbf{100.0} & $92.17\pm0.62$ & $68.67\pm1.25$ & $98.67\pm0.24$ & $95.5\pm0.4$ & $89.17\pm0.24$ \\
  \bottomrule
\end{tabular}
\end{table}

\begin{table}[t]
  \centering
  \caption{Goal-reaching success rates (\%) on Sokoban-Long; columns
are goal offsets measured in original environment steps.
$^{\dagger}$ denotes results reproduced by us.}
  \label{tab:results-sokoban}
  \scriptsize
  \setlength{\tabcolsep}{2.5pt}
  \renewcommand{\arraystretch}{1.05}
  \begin{tabular}{@{}lccc@{}}
    \toprule
    Method & 25 & 50 & 100 \\
    \midrule
    LeWM & $68.17\pm0.62$\textsuperscript{$\dagger$} & $41.50\pm0.71$\textsuperscript{$\dagger$} & $16.33\pm0.47$\textsuperscript{$\dagger$} \\
    Gemini & $69.67\pm0.85$\textsuperscript{$\dagger$} & $52.00\pm0.82$\textsuperscript{$\dagger$} & $35.00\pm0.41$\textsuperscript{$\dagger$} \\
    \midrule
    \multicolumn{4}{@{}l}{\textbf{Dual-WM}} \\
    \hspace{0.5em}From scratch & \textbf{\boldmath$87.83\pm0.62$} & \textbf{\boldmath$75.17\pm0.85$} & \textbf{\boldmath$52.33\pm0.94$} \\
    \hspace{0.5em}VFM adaptor & $83.67\pm0.24$ & $68.83\pm0.85$ & $48.00\pm0.41$ \\
    \hspace{0.5em}Low-level only & $87.00\pm0.00$ & $70.17\pm0.24$ & $45.83\pm0.85$ \\
    \bottomrule
  \end{tabular}
\end{table}

\begin{table}[t]
\centering
\caption{Goal-reaching success rates (\%) on PushT (left) and
Cube-Single (right); columns are goal offsets measured in original environment
steps. PushT carries the offset-$75$ column because HWM
reports its long-horizon result at offset $75$. On Cube-Single, Dual-WM (with
INTACT actor) adds the INTACT actor as an action prior for low-level
planning; all other Dual-WM rows use plain CEM.
$^{\dagger}$ denotes results reproduced by us.}
\label{tab:results-pusht}\label{tab:results-cube}
\scriptsize
\setlength{\tabcolsep}{1pt}
\setlength{\medmuskip}{2mu}
\renewcommand{\arraystretch}{1.05}
\begin{tabular}{@{}lccccccc@{}}
  \toprule
  \multirow{2}{*}{Method} & \multicolumn{4}{c}{PushT} & \multicolumn{3}{c}{Cube-Single} \\
  \cmidrule(lr){2-5} \cmidrule(lr){6-8}
  & 25 & 50 & 75 & 100 & 25 & 50 & 100 \\
  \midrule
  LeWM & 96.0 & $56.00\pm0.71$\textsuperscript{$\dagger$} & $45.67\pm0.62$\textsuperscript{$\dagger$}& $20.00\pm0.82$\textsuperscript{$\dagger$} & 74.0 & $51.33\pm1.03$\textsuperscript{$\dagger$} & $54.17\pm0.62$\textsuperscript{$\dagger$} \\
  DINO-WM & 74.0 & $55.33\pm0.62$\textsuperscript{$\dagger$} & $38.17\pm0.85$\textsuperscript{$\dagger$} & $12.00\pm0.71$\textsuperscript{$\dagger$} & 86.0 & $62.17\pm0.24$\textsuperscript{$\dagger$} & $62.33\pm0.47$\textsuperscript{$\dagger$} \\
  PLDM & 78 & -- & -- & -- & 65 & -- & -- \\
  HWM & 89 & 78 & 61 & $31.00\pm1.41$\textsuperscript{$\dagger$} & -- & -- & -- \\
  Hi-LeWM & $90.7\pm6.2$ & $42.0\pm6.8$ & $15.3\pm4.1$ & -- & -- & -- & -- \\
  Fast-LeWM & 98.0 & $41.00\pm1.41$\textsuperscript{$\dagger$} & $29.17\pm1.18$\textsuperscript{$\dagger$} & $9.50\pm0.82$\textsuperscript{$\dagger$} & 82 & $53.17\pm0.94$\textsuperscript{$\dagger$} & $66.33\pm1.31$\textsuperscript{$\dagger$} \\
  RC-aux & $90.8\pm3.3$ & -- & -- & -- & $76.0\pm7.5$ & $50.17\pm2.25$\textsuperscript{$\dagger$} & $65.33\pm1.89$\textsuperscript{$\dagger$} \\
  INTACT\textsuperscript{\textit{Pure CEM}} & $88.44\pm1.17$ & $33.67\pm1.89$\textsuperscript{$\dagger$} & $17.83\pm0.94$\textsuperscript{$\dagger$} & $5.67\pm1.65$\textsuperscript{$\dagger$} & $68.44\pm0.77$ & $53.50\pm1.08$\textsuperscript{$\dagger$} & $61.17\pm1.31$\textsuperscript{$\dagger$} \\
  INTACT\textsuperscript{\textit{Actor+CEM}} & $93.56\pm0.96$ & $47.83\pm0.85$\textsuperscript{$\dagger$} & $24.00\pm0.41$\textsuperscript{$\dagger$} & $11.50\pm0.71$\textsuperscript{$\dagger$} & $96.89\pm0.19$ & $57.17\pm1.03$\textsuperscript{$\dagger$} & $80.67\pm0.24$\textsuperscript{$\dagger$} \\
  JEPA-WM & $70.2\pm2.8$ & $38.17\pm1.25$\textsuperscript{$\dagger$} & $16.00\pm0.82$\textsuperscript{$\dagger$} & $5.33\pm1.18$\textsuperscript{$\dagger$} & -- & -- & -- \\
  VLWM & 94 & 60 & 20 & 12 & 74 & 54 & 50 \\
  Gemini controller & $22.67\pm0.24$\textsuperscript{$\dagger$} & $10.83\pm0.47$\textsuperscript{$\dagger$} & $8.00\pm1.41$\textsuperscript{$\dagger$} & $2.17\pm1.31$\textsuperscript{$\dagger$} & $56.17\pm1.18$\textsuperscript{$\dagger$} & $35.17\pm0.85$\textsuperscript{$\dagger$} & $30.17\pm1.03$\textsuperscript{$\dagger$} \\
  \midrule
  \multicolumn{8}{@{}l}{\textbf{Dual-WM}} \\
  \hspace{0.5em}From scratch & \textbf{\boldmath$98.00\pm0.71$} & \textbf{\boldmath$86.17\pm0.62$} & \textbf{\boldmath$62.00\pm0.41$} & \textbf{\boldmath$41.33\pm1.31$} & $74.00\pm0.41$ & $62.67\pm0.62$ & $67.83\pm0.47$ \\
  \hspace{0.5em}VFM adaptor & $67.17\pm0.24$ & $39.83\pm0.85$ & $23.33\pm0.94$ & $11.67\pm1.03$ & $79.33\pm0.85$ & \textbf{\boldmath$62.83\pm0.94$} & $68.00\pm0.71$ \\
  \hspace{0.5em}Low-level only & $97.83\pm1.03$ & $62.67\pm0.47$ & $53.33\pm0.85$ & $19.83\pm0.94$ & $74.33\pm1.03$ & $55.00\pm0.71$ & $61.50\pm1.08$ \\
  With INTACT actor & -- & -- & -- & -- & \textbf{\boldmath$97.33\pm0.24$} & $58.50\pm0.41$ & \textbf{\boldmath$82.50\pm0.71$} \\
  \bottomrule
\end{tabular}
\end{table}

The VFM-adaptor variant is strongest on Reacher, reaching $98.50\%$ at
offset $100$, but underperforms the from-scratch model on PushT.
The benefit of pretrained visual features is therefore task-dependent.
Cube-Single success is nonmonotone in dataset offset for several methods:
temporal separation along a recorded trajectory is not itself a measure
of minimum control difficulty. The discrete-action representation and
planner for Sokoban-Long are described in
Appendix~\ref{app:sokoban-implementation}.

\subsection{Actor-guided comparison on Cube-Single}
\label{app:actor-comparison}
Table~\ref{tab:actor-comparison} separates the action-proposal mechanism
from the world-model comparison. Without actor guidance, both methods use
CEM; with actor guidance, both use the INTACT actor. Dual-WM retains its
dual-latent dynamics and hierarchical planner in either setting.
At offset $100$, actor guidance improves both methods. At offset $50$,
it improves INTACT but lowers Dual-WM success from $62.67\%$ to $58.50\%$,
showing that its benefit depends on the evaluation setting.
These results complement the non-actor-guided aggregate in the main text.

\begin{table}[!htbp]
  \centering
  \caption{Cube-Single success (\%) grouped by method, with and without
  actor-guided proposals.
  Columns are goal offsets in environment steps. INTACT's offset-$25$
  results are quoted; the remaining entries are our evaluations.
  Bold marks a Dual-WM variant only when it attains the highest mean
  in the column.}
  \label{tab:actor-comparison}
  \small
  \begin{tabular}{@{}lccc@{}}
    \toprule
    Variant & 25 & 50 & 100 \\
    \midrule
    \multicolumn{4}{@{}l}{\textbf{INTACT}} \\
    \hspace{0.5em}Pure CEM & $68.44\pm0.77$ & $53.50\pm1.08$ & $61.17\pm1.31$ \\
    \hspace{0.5em}Actor+CEM & $96.89\pm0.19$ & $57.17\pm1.03$ & $80.67\pm0.24$ \\
    \midrule
    \multicolumn{4}{@{}l}{\textbf{Dual-WM}} \\
    \hspace{0.5em}Pure CEM & $74.00\pm0.41$ & \textbf{\boldmath$62.67\pm0.62$} & $67.83\pm0.47$ \\
    \hspace{0.5em}With INTACT actor & \textbf{\boldmath$97.33\pm0.24$} & $58.50\pm0.41$ & \textbf{\boldmath$82.50\pm0.71$} \\
    \bottomrule
  \end{tabular}
\end{table}

\FloatBarrier
\clearpage
\subsection{Success--time tradeoff on TwoRoom}
\label{app:planning-budget}

We evaluate four planning-budget settings for Dual-WM, its low-level-only
variant, LeWM, and HWM at goal offset $100$, with an execution budget of
$125$ environment steps. Each setting uses $200$ matched tasks per
planning seed ($0$, $1$, and $42$). Figure~\ref{fig:planning-budget}
reports mean success and standard deviation across these seeds.
Planning time per environment step is cumulative planning time divided
by the number of environment steps actually executed.
All methods run on an Intel Xeon E5-2680 v4 CPU and an NVIDIA RTX A6000
GPU in FP32 precision. Table~\ref{tab:planning-budget-sweep} lists the
CEM settings for each budget tier.

\begin{center}
  \includegraphics[width=0.90\linewidth]{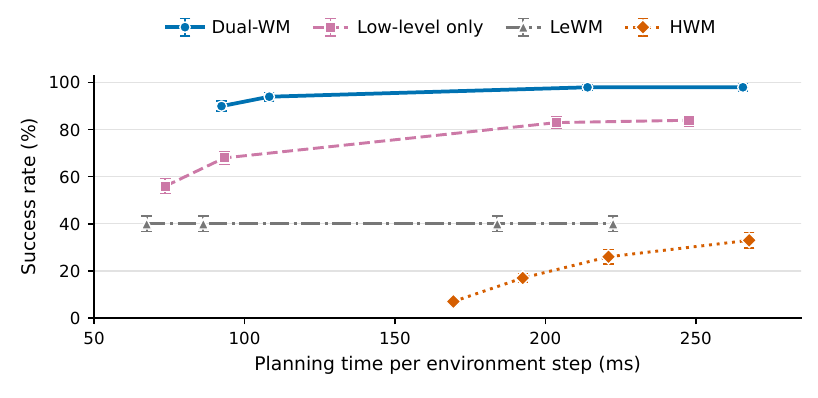}
  \captionof{figure}{Success--time tradeoff on TwoRoom at goal offset $100$ and an
  execution budget of $125$ environment steps. Each method has four
  measured planning-budget settings; lines connect successive settings.
  Error bars show $\pm$ one standard deviation across three planning seeds.}
  \label{fig:planning-budget}
\end{center}

Dual-WM achieves $90$--$98\%$ success over the measured range.
At nearly equal planning times, Dual-WM attains $90\%$ at $92.4$\,ms
and the low-level-only variant attains $68\%$ at $93.3$\,ms.
Even at its largest tested budget ($247.7$\,ms), the low-level-only
variant reaches $84\%$. LeWM remains at $40\%$ across its four settings,
while HWM improves from $7\%$ to $33\%$ as planning time increases.
These measurements support the benefit of dual-latent planning across
the tested range of planning costs.

\begin{center}
  \captionof{table}{CEM settings for the TwoRoom planning-budget sweep.
  Each entry lists candidates / elites / iterations. Tiers correspond
  to successive points on each method's curve in Figure~\ref{fig:planning-budget}.}
  \label{tab:planning-budget-sweep}
  \small
  \setlength{\tabcolsep}{7pt}
  \renewcommand{\arraystretch}{1.12}
  \begin{tabular}{@{}lcccc@{}}
    \toprule
    Method / level & Tier 1 & Tier 2 & Tier 3 & Tier 4 \\
    \midrule
    Dual-WM high & 30/10/7 & 40/10/8 & 80/10/16 & 100/10/20 \\
    Dual-WM low & 60/30/7 & 80/30/8 & 160/30/16 & 192/30/20 \\
    Low-level only & 60/30/7 & 96/30/9 & 192/30/20 & 256/30/25 \\
    LeWM & 60/30/7 & 96/30/9 & 192/30/20 & 256/30/25 \\
    HWM high & 30/10/7 & 40/10/8 & 80/10/12 & 80/10/12 \\
    HWM low & 30/30/3 & 35/30/3 & 40/30/3 & 60/30/3 \\
    \bottomrule
  \end{tabular}
\end{center}

\section{Qualitative execution across five tasks}
\label{app:qualitative-execution}

\begin{figure}[!htbp]
  \centering
  \includegraphics[width=\linewidth]{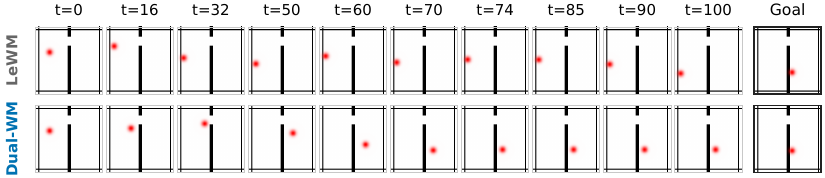}
  \caption{TwoRoom execution. Dual-WM navigates through the doorway and
  approaches the goal in the other room; LeWM remains on the starting
  side of the wall.}
  \label{fig:execution-tworoom}
\end{figure}

\begin{figure}[!htbp]
  \centering
  \includegraphics[width=\linewidth]{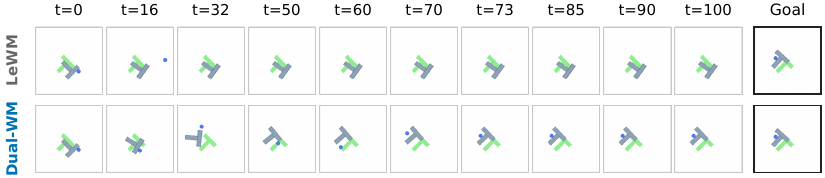}
  \caption{PushT execution. Dual-WM changes the block configuration toward
  the sampled goal, while LeWM leaves it near its initial configuration.
  The green shape is a fixed renderer reference; the task goal is the
  rightmost image.}
  \label{fig:execution-pusht}
\end{figure}

\begin{figure}[!htbp]
  \centering
  \includegraphics[width=\linewidth]{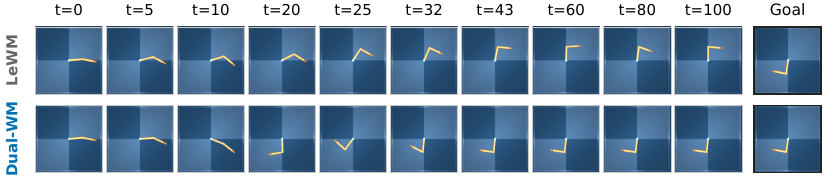}
  \caption{Reacher execution. Dual-WM brings the arm toward the goal
  configuration; LeWM moves the arm but retains a different configuration.}
  \label{fig:execution-reacher}
\end{figure}

\begin{figure}[!htbp]
  \centering
  \includegraphics[width=\linewidth]{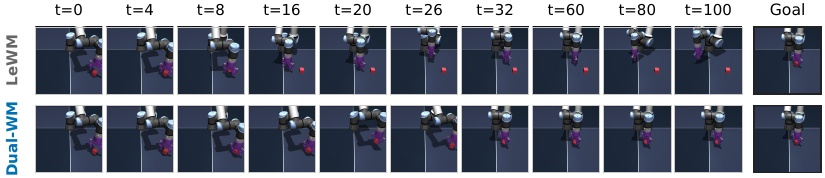}
  \caption{Cube-Single execution. Dual-WM brings the gripper and object
  toward the goal arrangement, whereas LeWM changes the gripper pose
  without completing the required object movement.}
  \label{fig:execution-cube}
\end{figure}

\begin{figure}[!htbp]
  \centering
  \includegraphics[width=\linewidth]{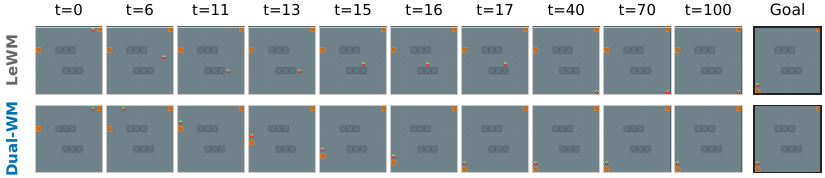}
  \caption{Sokoban-Long execution. Dual-WM navigates to a box and performs
  a sequence of aligned pushes toward the goal, while LeWM does not
  complete the required box movement. This supplemental held-out source
  pair lies outside the standard $200$-task sample.}
  \label{fig:execution-sokoban}
\end{figure}

Figures~\ref{fig:execution-tworoom}--\ref{fig:execution-sokoban}
show selected executions of LeWM and full from-scratch Dual-WM from
matched initial states and goals, with a common budget of $125$
environment steps. An offset of $100$ means that the initial and goal
frames are separated by $100$ environment steps within the same dataset
episode. Because the data-collection policies include substantial
randomness, these trajectories need not be direct routes to the sampled
goals, which can often be reached in fewer than $100$ steps.
Each row contains ten observation frames and the goal image;
timestamps count environment steps and are shared between methods.
After termination, the final recorded observation is repeated at later
timestamps for temporal alignment, including two intermediate timestamps
before $t=100$.

\paragraph{Long-range goal discrimination.}
The navigation and manipulation examples expose the need to distinguish
useful intermediate states from states that merely remain near the
starting configuration. In TwoRoom, progress requires a route through
the doorway before local convergence to the goal. In PushT and
Sokoban-Long, the agent must establish an appropriate contact or pushing
configuration before moving the object toward its target. These
behaviors complement the goal-ordering and distance-field results in
Sections~\ref{sec:concentration} and~\ref{sec:ablations}: a representation
that preserves long-range goal structure can guide intermediate choices,
while the low-level representation supports precise action refinement.

\paragraph{Maintaining useful predictions during execution.}
Completing these sequences also requires action-conditioned predictions
that remain informative across repeated planning updates. Dual-WM's
progress through distinct configurations provides a behavioral counterpart
to the recursive physical-state probes in Section~\ref{sec:physical-state}
and the LoRe scale ablation in Figure~\ref{fig:ablation-lore-scales}.
Taken together, the execution examples and quantitative diagnostics
support the complementary roles of long-range goal discrimination and
multi-step predictive consistency in goal-directed control.

\FloatBarrier
\section{Additional representation diagnostics}
\label{app:representation-fields}

\begin{figure}[!htbp]
  \centering
  \includegraphics[width=\linewidth]{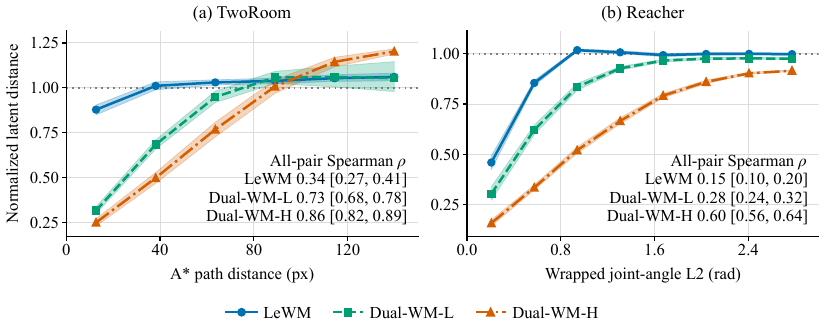}
  \caption{Latent-distance concentration on held-out state pairs. We plot
  dimension-normalized Euclidean latent distance against A* endpoint path
  distance in TwoRoom (left) and wrapped joint-angle distance in Reacher
  (right).  Curves are equal-width-bin means and shaded regions are 95\%
  episode-bootstrap confidence intervals.  The displayed task-relevant ranges
  are 0--150 px and 0--3 rad, respectively; annotated rank correlations use all
  held-out pairs.  Only bins with at least 100 valid pairs are displayed.  The
  dotted line marks the independent isotropic-Gaussian reference.}
  \label{fig:saturation}
\end{figure}

These diagnostics complement the held-out distance curves in
Section~\ref{sec:concentration}. TwoRoom visualizes obstacle-aware goal
geometry, Reacher examines configuration-space ordering, and PushT tests
goal discrimination in contact-rich manipulation.

\begin{figure}[!htbp]
  \centering
  \includegraphics[width=0.88\linewidth]{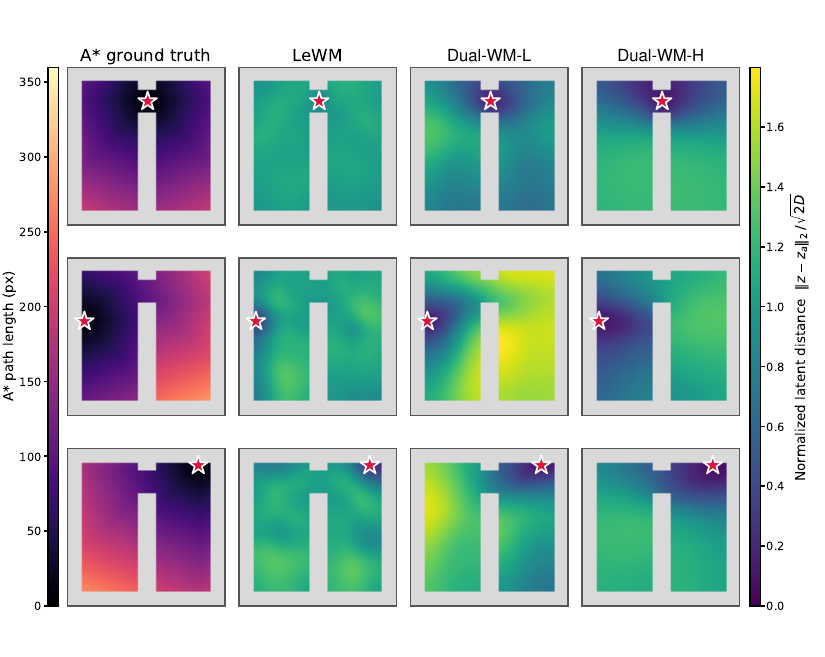}
  \caption{TwoRoom distance fields for three prespecified members of a
  geometry-only nine-anchor maximin set; stars mark the query states. Each row
  contains one A* reference followed by LeWM, Dual-WM-L and Dual-WM-H latent
  distances normalized by $\sqrt{2D}$. All A* panels share one path-length
  scale, and all model panels share one latent-distance scale. The wall-safe
  Gaussian bandwidth ($\sigma=10$ px) is selected by blocked spatial
  cross-validation over all three models and all nine anchors, without using
  A* similarity or visual appearance.}
  \label{fig:heatmap}
\par\medskip
  \includegraphics[width=0.88\linewidth]{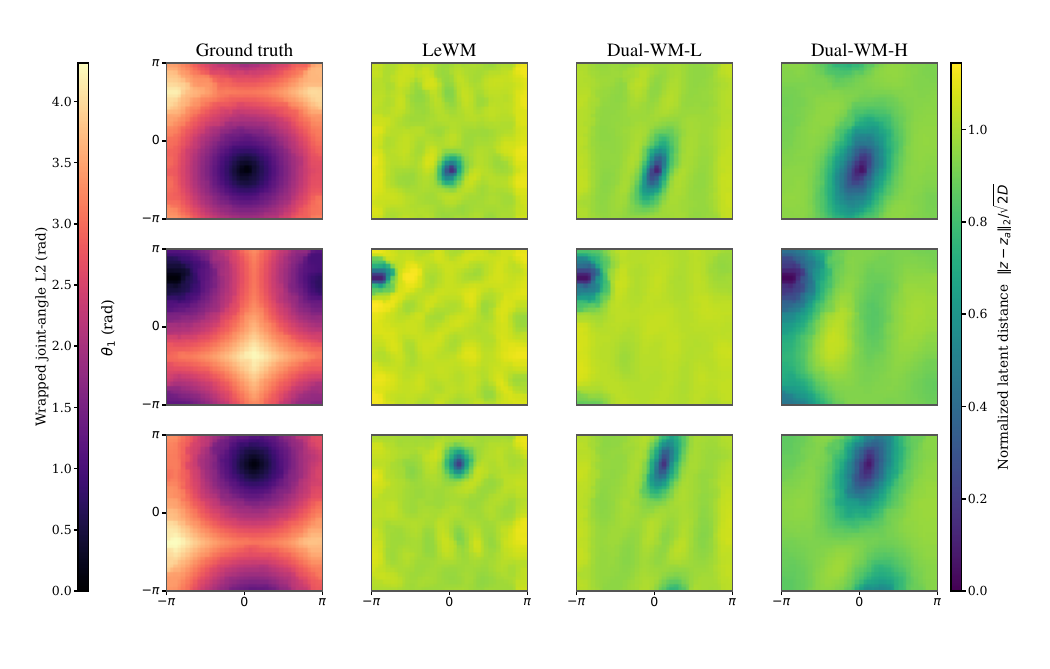}
  \caption{Reacher distance fields for three query states (rows). Columns show
  wrapped joint-angle distance, LeWM, Dual-WM-L, and Dual-WM-H. Model distances are
  normalized by $\sqrt{2D}$. Sparse grid cells are interpolated and lightly
  smoothed for visualization; all reported correlations are computed only on
  the 1,118 observed cells with at least ten samples.}
  \label{fig:reacher-heatmap}
\end{figure}

\begin{figure}[!htbp]
  \centering
  \includegraphics[width=0.68\linewidth]{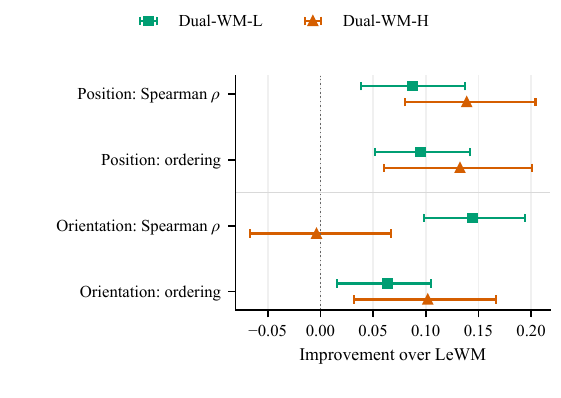}
  \caption{PushT goal-geometry improvements relative to LeWM on 128 held-out
  episodes. Dots are paired differences for Dual-WM-L and Dual-WM-H; error bars are
  95\% episode-cluster bootstrap confidence intervals (2,000 draws). Positive
  values favor Dual-WM. Correlations use physical goal distance, while ordering
  accuracy measures which state in a same-episode pair is closer to the goal;
  exact physical ties are excluded.}
  \label{fig:pusht-goal-geometry}
\end{figure}

\begin{figure}[!htb]
  \centering
  \includegraphics[width=0.92\linewidth]{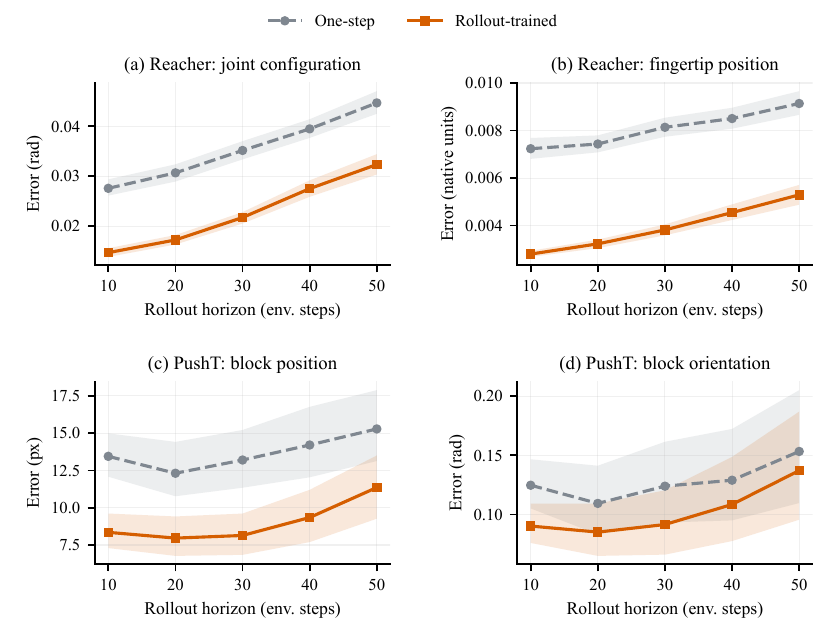}
  \caption{Physical state decoded from recursively predicted low-level
  latents.  A single horizon-pooled ridge probe is fitted separately for each
  checkpoint and physical target, selected on disjoint validation episodes,
  and evaluated on disjoint test episodes.  Curves report mean task-native
  error and shaded regions show 95\% episode-cluster bootstrap confidence
  intervals.  The rollout-trained checkpoint has lower point estimates at all
  horizons; PushT orientation has the widest uncertainty.}
  \label{fig:rollout-physical-probe}
\end{figure}

\paragraph{Physical-state probe protocol.}
For each checkpoint and physical target, a single ridge probe is fitted on
predicted latents pooled across all five horizons from training episodes.
Its regularization is selected on disjoint validation episodes, and it is
evaluated on disjoint test episodes. Probe fitting and evaluation therefore
use the same predicted-latent regime. Figure~\ref{fig:rollout-physical-probe}
reports the resulting task-native errors.

\section{Component-study protocols and sensitivity results}
\label{app:ablation-protocol}

\paragraph{Evaluation and units.}
The TwoRoom component studies use goal offsets of $25$, $50$, and $100$
original environment steps, with execution budgets of $50$, $75$, and
$125$ steps, respectively. Planning results average three planning seeds.
The component figures display means without uncertainty bars. The
latent-space and LoRe scale studies use the full hierarchical planner,
whereas the rollout-horizon and weighting studies use low-level-only
planning. The $69\%$ result at $N=5$ therefore evaluates the low-level
planner, not the full Dual-WM configuration used in the scale study.
Training
rollout lengths $N,M$, evaluation rollout lengths, and goal offsets have
different meanings: $N$ counts low-level model transitions, whereas goal
offsets count original environment steps.

\paragraph{Controlled variants.}
Within each ablation, settings other than the indicated factor remain at
their defaults. Shared-Identity passes the low-level endpoint latent
$z_t^L$ directly to the high-level predictor. Endpoint-MLP maps that endpoint through a three-layer MLP with
hidden width $2048$ to a distinct $192$-dimensional high-level state
used for both prediction and planning. Window-Concat uses the
same length-$k$ low-level latent window $W_t^L$ as Dual-WM, but replaces
$E_H$ with direct concatenation. Its high-level state is
$\operatorname{vec}(W_t^L)\in\mathbb R^{k d_L}$, and its high-level
predictor learns to predict the next macro-step window in this space.
All variants retain high-level planning followed by low-level refinement
and use the same frozen low-level checkpoint. Apart from the state interface
and its associated dimensions and architecture, training data, objectives,
training duration, and planning settings are identical. On TwoRoom,
$k=3$ and $d_L=192$: Shared-Identity, Endpoint-MLP, and Dual-WM use
$192$-dimensional high-level states, while Window-Concat uses $576$ dimensions. Shared-Identity and
Endpoint-MLP test endpoint-based interfaces, while Window-Concat controls
for the temporal context supplied to the learned window interface.
At offsets $50$ and $100$, Window-Concat achieves $69\%$ and $41\%$
success, compared with $100\%$ and $93\%$ for Dual-WM
(Figure~\ref{fig:ablation-latent-space}).

\paragraph{Goal-ordering protocol.}
For a goal $g$ and two observed states $s_1,s_2$, ordering is correct when
the sign of the difference between their latent goal distances agrees with
the sign of $d_{\mathrm{A*}}(s_1,g)-d_{\mathrm{A*}}(s_2,g)$.
Pairs with equal A* distance are excluded. The comparison uses the same
held-out state pairs and goals for all representations. Pairs are grouped
by $[d_{\mathrm{A*}}(s_1,g)+d_{\mathrm{A*}}(s_2,g)]/2$ into
$[0,30)$, $[30,60)$, $[60,90)$, $[90,120)$, and $[120,150]$ px bins.
Ordering is averaged within episodes and then across episodes.
Window-based representations receive observation context ending at each
candidate state; endpoint-based representations encode its final observation.
Dual-WM-H and Window-Concat use the same window. For a single goal image,
its low-level latent is repeated to form the goal window before encoding
or concatenation, respectively.
No future observation window is used for this diagnostic.
LeWM and Dual-WM-L are contextual references; Dual-WM-H, Endpoint-MLP,
Shared-Identity, and Window-Concat are the four planning representations
evaluated in the interface ablation.
In the nearest bin, Dual-WM-L reaches $99.8\%$ ordering accuracy versus
$94.2\%$ for Dual-WM-H and $93.5\%$ for Window-Concat.
In the farthest bin, Dual-WM-H retains $84.9\%$,
compared with $73.6\%$ for Endpoint-MLP, $56.6\%$ for Shared-Identity,
$55.6\%$ for Dual-WM-L, $52.4\%$ for LeWM, and $51.0\%$ for
Window-Concat. The widening gap between Dual-WM-H and Window-Concat
shows that equal window context does not yield equally informative
long-range goal rankings. Together with the planning results, this
supports learning a high-level geometry for distant subgoal evaluation
while retaining the low-level representation for local goal convergence.

\begin{figure}[!htb]
  \centering
  \includegraphics[width=0.85\linewidth]{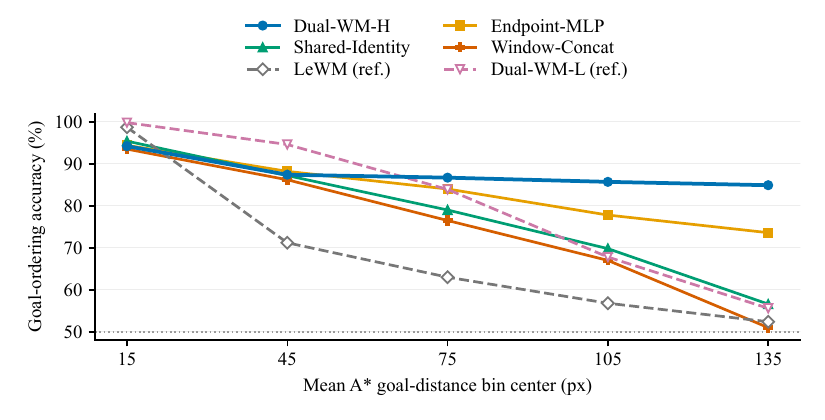}
  \caption{Goal ordering on TwoRoom. Accuracy compares latent and A* goal-distance rankings; bins use mean goal distance. Dual-WM-H and Window-Concat use identical observation windows, with learned encoding and direct concatenation, respectively. Dashed curves are contextual references. Corresponding planning success is shown in Figure~\ref{fig:ablation-latent-space}.}
  \label{fig:ablation-goal-ordering}
\end{figure}

\paragraph{LoRe scale activation.}
The one-step configuration retains both dynamics models and supervises
one transition at each level. The low-only and high-only configurations
activate LoRe at the indicated level while retaining one-step supervision
at the other. Both activates recursive supervision at both levels.
These training comparisons retain the hierarchical planner; they are
distinct from the low-level-only planning variant in the main result tables.

\paragraph{Low-level training horizon and probing.}
The horizon study uses low-level-only planning, varies
$N\in\{1,3,5,8,10\}$, and evaluates predicted
low-level states at horizons $1,2,3,5,7,10$. A position probe is fitted
on predicted latents pooled across horizons, with disjoint episodes for
fitting, validation, and evaluation. This measures how much position
information remains accessible after recursive prediction, rather than
the raw discrepancy between latent vectors from different models.
At evaluation horizon $10$, position error is $12.0$ px for $N=1$,
$5.7$ px for $N=5$, and $8.9$ px for $N=10$, consistent with the
planning advantage of a moderate training horizon in this study.

\begin{figure}[!htb]
  \centering
  \begin{minipage}[t]{0.48\linewidth}
    \centering
    \includegraphics[width=\linewidth]{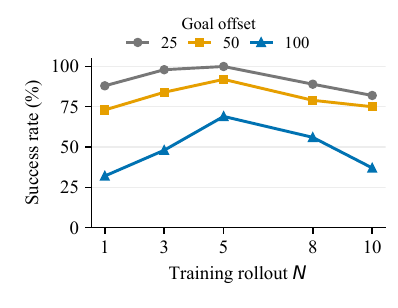}
  \end{minipage}\hfill
  \begin{minipage}[t]{0.48\linewidth}
    \centering
    \includegraphics[width=\linewidth]{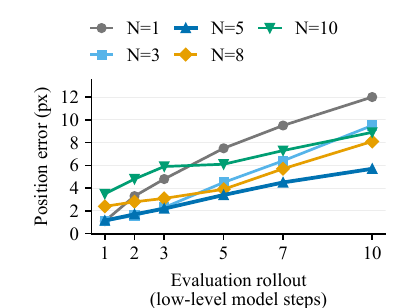}
  \end{minipage}
  \caption{Low-level training-horizon sensitivity on TwoRoom. Left: mean
  success using low-level-only planning at goal offsets $25$, $50$, and
  $100$ environment steps.
  Right: mean position-probe error over recursive low-level predictions.
  Both panels compare $N\in\{1,3,5,8,10\}$; the right horizontal axis is
  evaluation rollout length, not training length.}
  \label{fig:ablation-rollout-horizon}
  \label{fig:ablation-horizon-probe}
\end{figure}

\paragraph{Low-level rollout-weighting profiles.}
This study also uses low-level-only planning. For horizon $N$, uniform
weights are $w_h=1/N$, and linearly decreasing weights are
$w_h=2(N-h+1)/[N(N+1)]$, for $h=1,\ldots,N$.
Thus, at $N=5$, the linear profile is $(5,4,3,2,1)/15$.
The exponential profile is defined in Eq.~(\ref{eq:lore}).

\begin{figure}[!htb]
  \centering
  \includegraphics[width=0.58\linewidth]{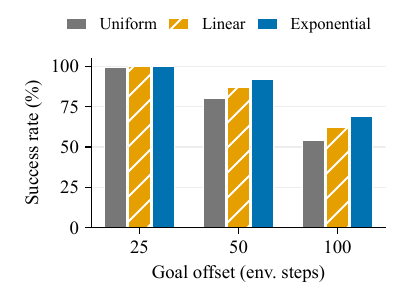}
  \caption{Rollout-weighting comparison on TwoRoom at training horizon
  $N=5$. Bars show mean success across low-level-only planning runs.
  The difference between profiles grows as the goal offset increases.}
  \label{fig:ablation-rollout-weighting}
\end{figure}

\begin{figure}[!htb]
  \centering
  \includegraphics[width=0.32\linewidth]{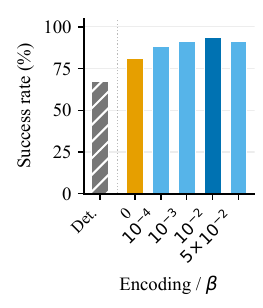}\hfill
  \includegraphics[width=0.32\linewidth]{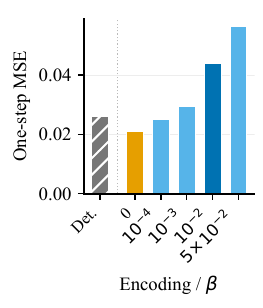}\hfill
  \includegraphics[width=0.32\linewidth]{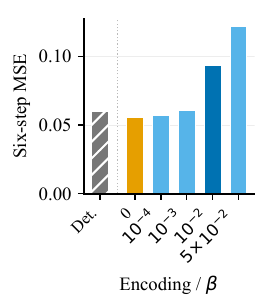}
  \caption{Macro-action prior shaping on TwoRoom. Left: mean success at
  goal offset $100$. Middle and right: mean one-step and six-step latent
  MSE. Det. denotes deterministic macro-action encoding; numeric labels
  denote stochastic encoding with the indicated MAPS coefficient,
  including $\beta=0$ without the KL penalty. All panels show the same
  six configurations. Raw MSE is measured in each learned latent space.}
  \label{fig:ablation-maps}
\end{figure}

\section{Sokoban-Long: Discrete-Action Implementation}
\label{app:sokoban-implementation}

Sokoban-Long extends the evaluation to discrete-action, game-like planning.
The world model uses vector-valued action inputs at both temporal scales,
while the low-level search distribution is selected according to the
environment's action space.

\paragraph{Action representation and preprocessing.}
The environment has a \texttt{Discrete(5)} action space:
$0$ is no-op, $1$ up, $2$ down, $3$ left, and $4$ right.
The frame skip is $f=1$, so one low-level model step corresponds to one
environment step. Actions are stored as five-dimensional one-hot vectors.
The low-level action encoder therefore has input width
$f\times\texttt{action\_dim}=5$, and the macro-action encoder $E_A$
receives a $k\times5$ window of one-hot actions. It encodes these $k$
discrete actions into a continuous macro-action posterior. This uses the
same encoder design and training objectives as the continuous-action
setting, with the appropriate action input width.
One-hot action columns are excluded from data normalization; only the
proprioceptive fields are normalized. In particular, statistics of the
five action columns are not applied to scalar environment action indices.

\paragraph{Categorical low-level search.}
For a discrete action space, the low-level planner selects categorical CEM
instead of Gaussian CEM. It maintains a categorical distribution over the
five actions at each planning step and samples candidate action sequences
using Gumbel-max. Each candidate is converted to a one-hot sequence before
latent rollout, matching the action representation used during training.
Candidates are ranked by the planning cost, and the distribution is refitted
to the empirical action frequencies of the elite sequences. Laplace
smoothing and exponential-moving-average updates prevent premature
concentration of the search distribution. High-level-guided refinement
uses the projected subgoal cost in Eq.~(\ref{eq:subgoal-refinement});
the final stage uses Eq.~(\ref{eq:goal-convergence}).

The low-level search uses $300$ candidates per iteration, $30$ iterations,
and $30$ elites, with planning horizon $H_L=12$. During high-level-guided
refinement, only the first action is executed before replanning from the
new observation. In the final stage, the categorical plan is cached and
reused over a four-environment-step execution horizon before replanning.
The selected one-hot actions are converted back to scalar action indices
for execution, without action denormalization.

For warm starts, unseeded sample rows use a sentinel value of $-1$.
Zero-filling these rows would instead designate action $0$ and bias the
search toward no-op. The sentinel marks missing warm-start entries and is
not an environment action.

\paragraph{Continuous high-level search.}
The macro-action $u\in\mathbb R^{16}$ remains continuous even though the
underlying actions are discrete. High-level planning uses Gaussian CEM
over the continuous macro-action space, with $100$ candidates per iteration,
$20$ iterations, and $10$ elites. Each candidate contains $H_H$ macro-actions;
this horizon corresponds to \texttt{u\_horizon} in the implementation.
We reserve $M$ for the high-level \emph{training} rollout length in the
paper. High-level search evaluates continuous macro-actions, while the
low-level planner searches discrete actions to realize its latent subgoals.
This separation allows the same hierarchical planning procedure to support
both action-space types through the low-level solver selection.

\section{Baseline descriptions and evaluation protocols}
\label{app:baselines}

Tables~\ref{tab:results-tworoom}--\ref{tab:results-cube} compare world-model
planners and an external visual controller. This appendix distinguishes
their representation learning, action-generation mechanisms, and evaluation
sources. Published entries retain their original evaluation conventions;
entries marked $\dagger$ are our reproduced planning evaluations.

\paragraph{Matched evaluation tasks and execution protocol.}
For every task and goal offset evaluated by us, each planning seed
($0$, $1$, or $42$) determines a set of $200$ goal-reaching tasks.
All methods receive exactly the same episode IDs, start states, and goals
within that seed. Each seed resamples the task set and fixes the CEM
search random stream, yielding repeatable per-task outcomes for a given
planner. The reported standard deviation captures variation across these
seeded evaluations in both task sampling and search randomness.

For a given task and offset, all methods evaluated by us use the same
maximum number of environment steps, success criteria, and
early-termination rules. This shared protocol applies to Dual-WM and
all reproduced baselines. Results quoted from prior publications retain
their original protocols; the search configurations used in our evaluations
are specified below.

\paragraph{Model sources for reproduced results.}
For entries marked $\dagger$, DINO-WM and HWM are trained on the same
datasets as Dual-WM using their authors' released code. LeWM, Fast-LeWM,
RC-aux, INTACT, and JEPA-WM use the authors' released checkpoints and
are evaluated on our matched test tasks. JEPA-WM's offset-$25$ result
is quoted, whereas its offset-$50$, $75$, and $100$ results are our
evaluations of the released model. Our CEM-based baseline evaluations
use $300$ candidates, $30$ iterations, and $30$ elites for low-level
action search. High-level macro-action search uses $100$ candidates,
$20$ iterations, and $10$ elites. Results without the reproduction
marker retain the reporting convention of the original publication,
apart from our own Dual-WM results.

\paragraph{Latent dynamics baselines.}
\emph{LeWM}~\citep{maes2026leworldmodel} trains an encoder and a predictor jointly
with an anti-collapse regularizer, and plans in the resulting latent space.
\emph{DINO-WM}~\citep{zhou2024dinowm} predicts the features of a pretrained DINO
encoder instead of reconstructing pixels.
\emph{PLDM}~\citep{sobal2025pldm} learns dynamics from reward-free offline
trajectories and is reported at offset~$25$.
\emph{JEPA-WM}~\citep{terver2025whatdrives} is reported on PushT at offsets $25$,
$50$, $75$ and $100$.

\paragraph{Hierarchical and multi-step baselines.}
\emph{HWM}~\citep{zhang2026hwm} learns two temporal scales inside one shared latent
space. We reproduce its TwoRoom results at offsets $25$, $50$, and $100$,
and its PushT result at offset $100$. The PushT results at offsets $25$,
$50$, and $75$ are quoted from the original publication; the offset-$75$
result motivates the extra column.
\emph{Hi-LeWM}~\citep{caselli2026hilewm} adds hierarchical planning on top of LeWM
and is reported on PushT at offsets $25$, $50$ and $75$.
\emph{Fast-LeWM}~\citep{gao2026fastlewm} predicts action-prefix outcomes in parallel
from an observed anchor.
\emph{RC-aux}~\citep{li2026rcaux} keeps recursive dynamics and supervises them
open-loop with a horizon-weighted auxiliary objective.
\emph{VLWM}~\citep{du2026vlwm} predicts the outcome of variable-length action
sequences and is reported on TwoRoom, PushT and Cube-Single.

\paragraph{Actor-guided planning.}
INTACT~\citep{sun2026intact} learns an intent-to-action interface from
action-labeled trajectories. We distinguish Pure CEM, which searches without
actor-guided proposals, from Actor+CEM, which uses the learned action
interface to guide search. The offset-$25$ entries are quoted from the
original paper and retain its reporting convention. The offset-$50$,
offset-$75$, and offset-$100$ entries are our reproduced evaluations.
Dual-WM with the INTACT actor is a separate extension on Cube-Single:
it adds actor-guided low-level proposals while retaining dual-latent
dynamics and hierarchical planning. It is excluded from the from-scratch
cross-task aggregate.

\paragraph{External visual controller.}
We also evaluate a Gemini 3.8 Flash controller as a closed-loop reference.
The controller receives the current observation, goal image, and action
space, and produces an action chunk. We execute five environment steps
before requesting a new plan. It provides an external visual-control
reference alongside the learned world-model planners.
The discrete-action implementation and execution schedule of Dual-WM are
described in Appendix~\ref{app:sokoban-implementation}.

\paragraph{Cross-task aggregation.}
For goal offset $m$, let $S_{t,b}(m)$ be the reported mean success of method
$b$ on task $t$. We compute
\[
  \overline S_b(m)=\frac{1}{5}\sum_{t=1}^{5}S_{t,b}(m),\qquad
  \overline S_{\mathrm{best}}(m)=
  \frac{1}{5}\sum_{t=1}^{5}\max_{b\in\mathcal B_t(m)}S_{t,b}(m),
\]
where $\mathcal B_t(m)$ contains non-Dual-WM methods with a reported
result for that task and offset and without actor-guided proposals.
The same eligibility rule applies on every task: INTACT uses Pure CEM,
and Actor+CEM is excluded. The external Gemini controller is retained.
The selection uses available reported results. At offset $50$, the
selected methods are Gemini on TwoRoom and Sokoban-Long, INTACT (Pure CEM)
on Reacher, HWM on PushT, and DINO-WM on Cube-Single. At offset $100$,
they are Gemini on TwoRoom, Reacher, and Sokoban-Long, HWM on PushT,
and Fast-LeWM on Cube-Single. The resulting means are $75.9\%$ and
$61.4\%$. All Dual-WM aggregate values use the from-scratch variant.
Aggregates are computed before rounding to one decimal place.

For completeness, allowing actor-guided competitors in the task-wise
selection raises these aggregates to $76.9\%$ and $65.9\%$.
At offset $50$, INTACT (Actor+CEM) replaces Pure CEM on Reacher.
At offset $100$, it replaces Gemini on Reacher and Fast-LeWM on
Cube-Single. This broader comparison is distinct from the proposal
grouping used in the main table.

\FloatBarrier
\Needspace{0.9\textheight}
\section{Additional TwoRoom distance fields}
\label{app:tworoom-fields}

The six query states not displayed in Figure~\ref{fig:heatmap} are collected
below under the same smoothing and shared color-scale protocol. Each star
marks the corresponding query state.
\begin{figure}[!htbp]
  \centering
  \includegraphics[height=0.74\textheight,keepaspectratio]{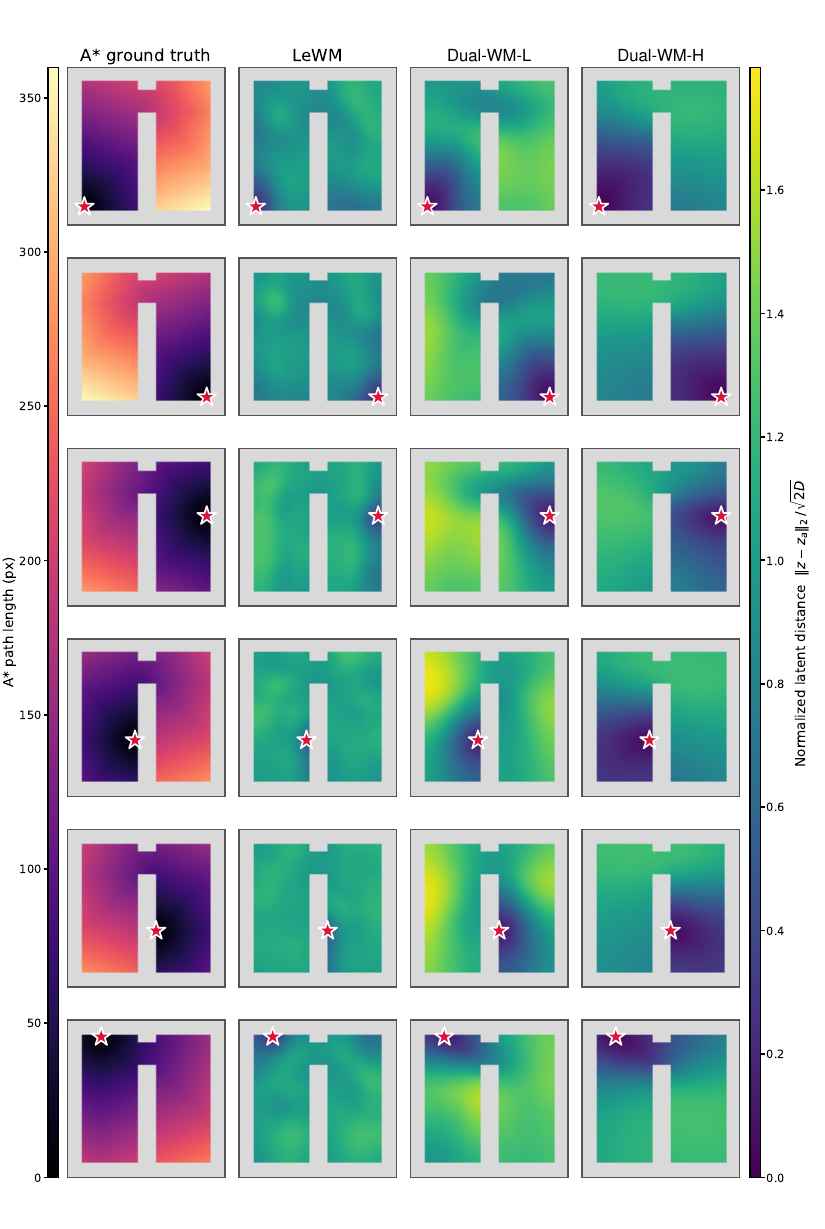}
  \caption{Additional TwoRoom distance fields for the six remaining
  query states. Each row contains one A* reference followed by LeWM, Dual-WM-L
  and Dual-WM-H. The A* column and the three model columns use the same respective
  shared scales as Figure~\ref{fig:heatmap}.}
  \label{fig:heatmap-appendix}
\end{figure}

\end{document}